\documentclass[11pt]{article}

\usepackage[margin=1in]{geometry}
\usepackage[utf8]{inputenc}
\usepackage[T1]{fontenc}
\usepackage{lmodern}
\usepackage{microtype}
\usepackage{graphicx}
\usepackage{booktabs}
\usepackage{array}
\usepackage{amsmath}
\usepackage{xcolor}
\usepackage[hyphens]{url}
\usepackage[numbers,sort&compress]{natbib}
\usepackage{tikz}
\usetikzlibrary{positioning,arrows.meta}
\usepackage[colorlinks=true,linkcolor=blue!60!black,citecolor=blue!60!black,urlcolor=blue!60!black]{hyperref}

\newcommand{\system}{Mingbird}
\newcommand{\lrab}{LRAB}

\title{Mingbird: A Local-First Agent Harness Enabling\\ Small Open Models to Complete Real Tasks}

\author{Hao Wang\textsuperscript{$\dagger\,\ast$}\\
School of Mathematics and Physics, University of Science and Technology
Beijing\\
Beijing, China\\
\texttt{wanghao994916959@163.com}\thanks{$\dagger$ These authors
contributed equally. $\ast$ Co-corresponding authors. H.\ Wang's
contribution is independent work, not affiliated with any employer of the
second author. Code, benchmark
protocol, scoring code, and all 288 per-cell results are available in the
project repository (Apache-2.0),
\url{https://github.com/Mingbird/Mingbird-agent}. The harness is developed
under the GitHub handle \texttt{Gustor-Wang}.}\and
Ting Huang\textsuperscript{$\dagger\,\ast$}\\
Honor Device Co., Ltd.\\
Beijing, China\\
\texttt{huangting6@honor.com}}
\date{October 2026}

\begin{document}
\maketitle

\begin{abstract}
Small open-weight models (2--9B) run on ordinary laptops, but under
cloud-scale agent harnesses they rarely complete real tasks: tool prefill
overflows the context, self-correction diverges, tool demonstrations
loop, and tasks are silently abandoned. We present evidence, from a
controlled single-machine comparison and one third-party benchmark, that
a substantial share of these failures is attributable to the harness
rather than the model. We introduce \system{}, a local-first agent
harness for Windows and Ollama whose ten mechanisms compensate
point-by-point for small-model failure forms, three of them
representative: a byte-level net-zero prefill budget, a finish gate that
re-reads the task before accepting completion, and signature-level loop
detection. On \lrab{}, a controlled comparison holding machine, models,
budgets, and scoring fixed (4 harnesses $\times$ 4 open models
(2B--35B) $\times$ 18 real tasks, deterministic artifact scoring),
\system{} reaches 0.886 overall against 0.631 (goose), 0.479 (opencode),
and 0.405 (agent-mini), with all 288 cells published; on $\tau^2$-bench
(278 tasks, three arms, one protocol) it totals 0.856 against 0.791 and
0.737; and a frontier-model probe on the same 18 tasks spans 0.997 to
0.478 across harnesses, with well-formed scaffolds staying within 0.072 of
each other. A leave-one-mechanism-out ablation is reported as
directional only: same-night replications of the same arm move its mean by
up to 0.069, the size of every nominal single-trial delta, and the one
batch-matched comparison (full mechanism stack versus text re-read alone)
gives the executable completion guards a paired $+0.10$ across three
replications. The
evidence carries stated limits: a self-built benchmark, a single
machine, and single-trial scoring.
\end{abstract}

\section{Introduction}
\label{sec:intro}

Open-weight language models in the 2--9B parameter range have crossed a
practical threshold. On a laptop with an integrated GPU and 16--32\,GB of
shared memory they answer, summarize, translate, and explain at a level that
makes them useful daily tools
\citep{slmsurvey,smollm2,gemma,qwen25}. Local runtimes such as
Ollama and llama.cpp made deployment a one-command operation
\citep{ollama,llamacpp}. The prevailing assumption is that such models cannot
serve as agents --- systems that autonomously execute multi-step,
tool-mediated tasks: writing code that passes its tests, organizing files,
conducting research with real search, or driving a domain database
\citep{tinyagent,lostmiddle}. A recent position paper argues the
opposite for agentic workloads \citep{belcak2025slmagents}; our results
support that view, with one qualification: the limiting component is usually
the harness rather than the model.

Our observations conflict with this assumption in a specific way: small models
generally produce the correct next action, but cannot reliably emit and
execute every step unaided, inside a harness designed for a much larger
model. When we drove a 2B model with a cloud-scale agent framework, we
observed the same failure forms repeatedly: the static tool prefill alone
consumed a large fraction of the context (long-context degradation is
well documented \citep{lostmiddle}); the model could not self-correct
after a failed tool call \citep{kamoi2024selfcorrect,huang2023selfcorrect}; it looped indefinitely on tool demonstrations; and
it sometimes declared a task finished when the artifacts did not exist, or
gave up halfway without saying so. These are not failures of capability.
They are failures of the scaffold surrounding the model, and we treat them as
engineering defects with corresponding engineering remedies.

This paper makes that argument concrete. We describe \system{}, a
local-first agent harness for Windows and Ollama, designed from the start for
2--9B models (upper bound tested at the 35B tier; Table~\ref{tab:models}).
\system{} is a GUI application and a CLI over one agent loop; it adds no
inference machinery of its own and talks to whatever model Ollama serves. Its
design consists of ten core mechanisms (M1--M10) plus a five-ring safety
model (M14, v1.6.0; M11--M13 are product-level facilities,
Section~\ref{sec:mechanisms}), each mapped to a failure form we first observed
and then measured.

Measuring harness effects requires holding everything else constant, so we
built \lrab{} (Local Real-task Agent Benchmark) as a controlled experiment:
4 harnesses (\system{}, goose \citep{goose}, agent-mini \citep{agentmini},
and opencode \citep{opencode}, all stock and unpatched) $\times$ 4 open
models (2B, 4B, 12B, 35B; Table~\ref{tab:models}) $\times$ 18 real tasks,
288 cells
in total, on one machine, one backend, uniform time budgets, and deterministic
artifact-based scoring. Every one of the 288 cells is published as raw data;
every aggregate from the unified-protocol campaigns is directly recomputable
from the rows of a published CSV file (the legacy thinking-on sensitivity
numbers, retained for comparison in Section~\ref{sec:tau2}, are
aggregate-level only). Because \lrab{} was designed by us, we also
validate externally: on $\tau^2$-bench \citep{tau2bench}, an independent
benchmark we did not design, we ran all three domains (retail, airline,
telecom) under one uniform protocol
(Section~\ref{sec:tau2}). Over the 278 scored tasks (114 retail, 50 airline,
114 telecom), \system{} totals 0.856 against 0.791 for the benchmark's
native agent and 0.737 for opencode; the goose arm is deferred for
throughput (Section~\ref{sec:tau2}).

\paragraph{Contributions.}
\begin{enumerate}
  \item \textbf{A mechanism set for small-model harnesses.} Ten core design
  patterns plus a five-ring safety model (Section~\ref{sec:harness}), each
  documented in the form
  \emph{failure form $\rightarrow$ mechanism $\rightarrow$ design trade-off},
  implemented in an open-source, Apache-2.0 harness and covered by 441 tests
  at the documented v1.6.0 release (461 at the v1.8.2 tag current at
  writing, Section~\ref{app:repro}) and a byte-level prefill regression gate; the
  safety model is additionally evaluated against a 54-case adversarial
  probe set on both small-tier models, one run per case
  (Section~\ref{sec:m14}).
  \item \textbf{A controlled comparison methodology.} \lrab{}
  (Section~\ref{sec:lrab}) holds model, machine, backend, budgets, and scoring
  constant and varies only the harness; we frame the benchmark explicitly as
  a controlled experiment rather than a leaderboard, with the task set, the
  scoring code, and the complete data release published alongside it.
  \item \textbf{288-cell results.} \system{} 0.886 overall vs.\ 0.631 (goose),
  0.479 (opencode), 0.405 (agent-mini); per-model, \system{} is the only
  harness without a small-model cliff (2B: 0.821 vs.\ 0.017--0.271; a second
  small-tier family run through all four harnesses reproduces the pattern at
  0.779 vs.\ 0.017--0.239), and it
  leads the long-horizon segment at 0.827 (single draw per cell;
  Section~\ref{sec:results}).
  \item \textbf{External validation.} On $\tau^2$-bench, all three domains
  (retail, airline, telecom) run under one protocol:
  over the 278 scored tasks, \system{} totals 0.856 vs.\ 0.791 (native
  agent) and 0.737 (opencode), derived, with the near-saturated
  telecom domain reported as such and the fourth CLI arm (goose) deferred
  for throughput (Section~\ref{sec:tau2}).
  \item \textbf{Mechanism-level evidence.} A leave-one-mechanism-out
  ablation at 2B ($n=18$ per arm, same code base) covers three mechanisms
  (M1, the two halves of M3, and M4) across four variant arms and yields
  directional marginal effects (removing the finish gate reduces the total
  by 0.098; the flat prefill is near-neutral on this task set at 2B;
  Section~\ref{sec:ablation}).
\end{enumerate}

We are explicit about what this paper is not. It proposes no new algorithm;
the model, the backend, and the prompts are ordinary. The contribution is
systems engineering: identifying which harness behaviors break small models
and showing that fixing them moves measured outcomes by large margins.

\section{Background and Related Work}
\label{sec:related}

\subsection{Agent paradigms}
The dominant agent recipe descends from ReAct \citep{react}: the model
interleaves free-form reasoning with tool calls in a loop, and the harness
executes each call and appends the observation. Toolformer \citep{toolformer}
demonstrated that models can learn when to call APIs; Gorilla
\citep{gorilla} fine-tunes specifically for accurate API invocation and
motivated the function-calling interfaces now exposed by most inference
providers. Around this core, a body of work improves the \emph{model's}
behavior inside the loop: Reflexion \citep{reflexion} stores verbal
self-critiques across attempts, and Self-Refine \citep{selfrefine} iterates on
its own outputs. These techniques presuppose exactly what small models lack,
namely the ability to produce a usable critique of one's own attempt.
\system{} therefore concentrates on the harness side: it does not ask the
model to self-improve in the abstract; it changes what the model sees, when it
is allowed to stop, and what happens after each malformed step. Anthropic's
engineering guidance \citep{anthropic_agents} distinguishes workflows
(predefined code paths) from agents (the model directs its own process);
\system{} is an agent in that sense, but with harness-enforced guardrails at
the points where small models demonstrably fail.

\subsection{Agent benchmarks}
AgentBench \citep{agentbench} and AgentBoard \citep{agentboard} evaluate
multi-turn agents across many environments, the latter with analytical
progress metrics rather than final success only. WebArena \citep{webarena}
builds a reproducible web environment; SWE-bench \citep{swebench} scores real
GitHub issue resolution; GAIA \citep{gaia} scores general assistant questions
that are easy for humans. Closest to our external validation are $\tau$-bench
\citep{taubench} and $\tau^2$-bench \citep{tau2bench}: a user simulator
converses with an agent that must operate domain tools under policy, and
scoring verifies the final database state, so an agent that bypasses its tool
stream cannot score. \lrab{} differs from all of these in intent: it is not
meant to rank models but to isolate harness effects under a fully published
protocol, which is why it uses one machine, one backend, and per-cell
deterministic artifact scoring.

\subsection{Local and small-model inference}
llama.cpp \citep{llamacpp} established efficient CPU/GPU inference for
quantized open models, and Ollama \citep{ollama} packages it with model
management and a serving API. Surveys of small language models
\citep{slmsurvey} document the on-device latency and memory envelope, and
data-centric efforts such as SmolLM2 \citep{smollm2} and the Gemma and Qwen
families \citep{gemma,qwen3tr} push capability per parameter. This line of work
improves the model side of the local stack. The harness side, however, has
largely been ported from cloud practice: lightweight libraries such as
smolagents \citep{smolagents} and Open Interpreter \citep{openinterpreter}
make small models drivable with minimal scaffolding; function-calling at
the edge has been explored directly \citep{tinyagent}; and the full harnesses we
test against (goose \citep{goose}, opencode \citep{opencode}, agent-mini
\citep{agentmini}) are designed around large-context cloud models and
Unix-family shells. \system{} starts from the opposite end: 2--9B models, a
Windows host, an integrated GPU, and tasks that mix code, files, and research.

\subsection{Engineering systems}
SWE-agent \citep{sweagent} showed that the agent--computer interface (the
shape of the tools an agent is given) changes measured performance on
SWE-bench; our results agree and extend the observation to model-scale
effects. OpenHands \citep{openhands} is a general platform for software-development
agents. Cline \citep{cline} and aider \citep{aider} embed agents in editors
and git workflows. These systems target strong models (cloud APIs or large
local checkpoints), assume a POSIX-flavored command environment, and optimize
for coding tasks. \system{} is complementary: it targets the
under-served regime where the model is small, the machine is a consumer
laptop, and the platform is Windows, and it treats harness mechanisms for
small-model failure forms as the primary design object.

\paragraph{Harness and scaffold effects.}
That the harness, not only the model, drives measured agent performance is
an active 2026 result, not our discovery. Zhang et al.\
\citep{harnessdisclosure} name and test the binding-constraint thesis ---
harness configuration dominates model choice as a source of performance
variance --- and call for mandatory harness disclosure; Starace
\citep{scaffoldgaia} pre-registers a three-scaffold, five-model comparison
on GAIA and finds within-model scaffold effects up to 28 points; the CAIS
Arena benchmark \citep{caisarena} pins a single model across six agent
frameworks on one MCP tool server. Our contribution is complementary and
at a tier none of these cover: we instantiate the thesis with 2--9B local
weights on Windows/Ollama, and decompose it at the mechanism level --- a
reusable inventory of small-model failure forms with the harness mechanism
that addresses each, and measured per-mechanism costs and deltas. For loop
detection specifically, OpenHands' StuckDetector \citep{openhands}
normalizes events and counts repeated action--observation ($\geq$4),
action--error ($\geq$3), monologue, and alternation patterns with
intervention nudges; our M4 follows that detection paradigm and adds
signature normalization over tool name plus arguments, a tiered escalation
ladder, and a CI-enforced regression tie. What we claim as new at this
scale is narrower: a net-zero prefill budget enforced byte-level in CI,
and a completion gate that re-reads the task text before accepting
``done'' --- both aimed at the 2--9B tier where, per our measurements,
prefill crowding and false finishes are the dominant failure forms. A
complementary line treats the harness itself as the design object: Agentless
\citep{agentless} shows that a deliberately narrow scaffold can outperform
agentic ones on software-engineering tasks; ADAS \citep{adas}, AFlow
\citep{aflow}, GPTSwarm \citep{gptswarm}, and DSPy \citep{dspy} search,
compile, or optimize harness structure automatically; and ToolLLM
\citep{toolllm} routes over large API spaces with a learned retriever. Our
category router is a static, zero-training counterpart of that retrieval
idea, and our mechanism set is hand-designed rather than searched --- a
choice Section~\ref{sec:ablation} quantifies mechanism by mechanism.

\section{The \system{} Harness}
\label{sec:harness}

\subsection{Architecture overview}

\begin{figure}[htbp]
\centering
\begin{tikzpicture}[
  font=\small,
  box/.style={draw, rounded corners=2pt, align=center, inner sep=5pt,
              minimum height=9mm},
  wide/.style={box, text width=52mm},
  mech/.style={box, text width=30mm, font=\footnotesize},
  side/.style={box, text width=34mm, font=\footnotesize, dashed},
  arr/.style={-{Latex[length=2mm]}, thick},
  node distance=4mm and 6mm
]
\node[wide] (frontends) {Frontends\\ GUI (ttkbootstrap, bilingual) / CLI};
\node[wide, below=of frontends] (loop) {\textbf{Agent loop}\\
  prefill router (M1) $\cdot$ chat/task layer (M2)\\
  finish gate (M3) $\cdot$ anti-loop monitor (M4)\\
  format rescue (M5)};
\node[wide, below=of loop] (exec) {\textbf{Tool executor}\\
  shell $\cdot$ file operations (.bak, M6)\\
  MCP tools (flat injection, M1)\\
  skills (progressive disclosure, M9)};
\node[wide, below=of exec] (backend) {\textbf{Ollama backend}\\
  local open models, 2B--35B\\ ctx 128K default (256K max)};
\node[side, right=of loop] (guards) {\textbf{Guardrails (M8)}\\ path boundary
  $\cdot$ sensitive-path gate\\ GUI confirmation for out-of-bounds\\
  default-deny sub-agents};
\node[side, right=of exec] (resil) {\textbf{Resilience (M7)}\\
  encoding-tolerant reads\\ empty-turn hard reset\\ kill-resume checkpoints};
\node[side, right=of backend] (quality) {\textbf{Quality base (M10)}\\
  441 tests $\cdot$ CI\\ prefill net-zero regression\\ assertion};
\draw[arr] (frontends) -- (loop);
\draw[arr] (loop) -- (exec);
\draw[arr] (exec) -- (backend);
\draw[arr, dashed] (guards.west) -- (loop.east);
\draw[arr, dashed] (resil.west) -- (exec.east);
\draw[arr, dashed] (quality.west) -- (backend.east);
\end{tikzpicture}
\caption{\system{} architecture. Mechanism identifiers (M1--M10) refer
to Section~\ref{sec:mechanisms}; the five-ring safety model (M14) is a
policy layer over these components and is not drawn. All components run
locally on the user's machine; the harness talks to the Ollama API and,
when the user explicitly configures one, an external search endpoint. In
the shipped default configuration, no other outbound traffic exists; the
evaluation infrastructure of Sections~\ref{sec:protocol}
and~\ref{sec:tau2} adds a local protocol proxy and, for the $\tau^2$
comparison, a cloud-hosted user simulator and judge.}
\label{fig:arch}
\end{figure}
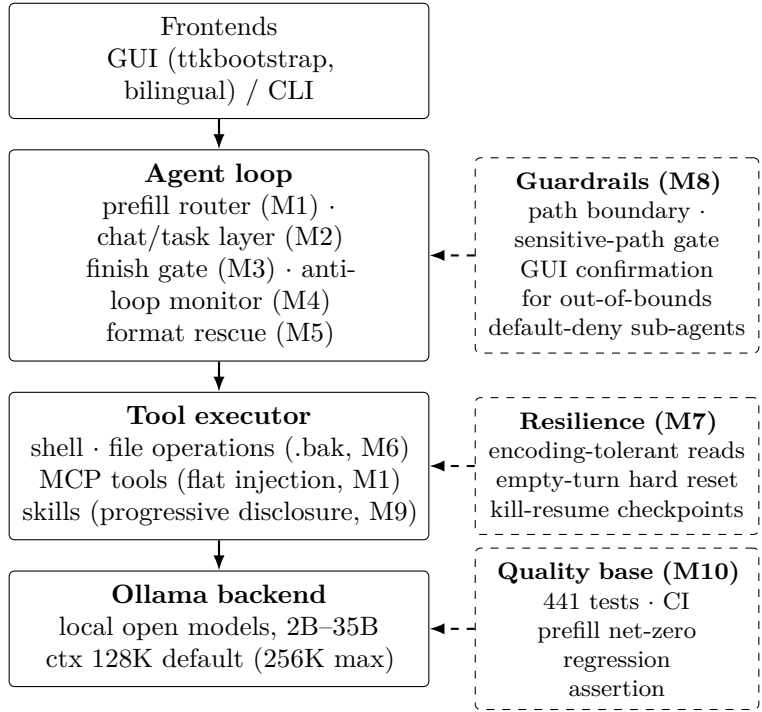

\system{} is a local-first agent harness: everything (model, tools,
speech recognition, memory) runs on the user's machine (the only optional
outbound endpoint is a user-configured search provider), and the harness
adds no inference overhead of its own. This section describes the v1.6.0
release; all measurements reported in this paper were taken on v1.5.0, and
the two releases differ by 23 static-prefill tokens (774 vs.\ 797;
Appendix~\ref{app:repro}). Figure~\ref{fig:arch} shows the
architecture. A GUI (bilingual, built on ttkbootstrap) and a CLI front the
same agent loop. Per turn, the loop assembles a \emph{flat prefill} through a
task-domain router (M1), classifies the request as chat or task (M2),
executes tool calls through a sandboxed executor with edit safety (M6) and
security boundaries (M8), monitors loop signatures (M4), and enforces a
finish gate before any completion claim leaves the system (M3). A resilience
layer (M7) absorbs encoding faults, empty turns, and process kills via
checkpoints. The backend is whatever Ollama serves, with automatic model
detection and a default 128K context (up to 256K).

Two quantities bound the resource footprint. The factory prefill
(everything static the model sees before any task content) is 797 tokens in
the current release, pinned by a unit test rather than allowed to drift. The
quality base is 441 tests at the documented v1.6.0 (461 at the v1.8.2
tag current at writing; repository test counts and their unit/integration
split are stated in the README), including a regression assertion
that any change to the codebase must keep the static prefill at exactly zero
net growth, checked byte-for-byte. The rest of this section expands each
mechanism.

\subsection{Mechanisms}
\label{sec:mechanisms}

Table~\ref{tab:mech} maps each mechanism to the failure form it addresses.
The subsections give the details and the trade-offs.

\begin{table}[t]
\centering
\caption{\system{} mechanisms (M1--M10, plus the v1.6.0 five-ring safety
model M14) and the small-model failure forms
they address. Each mechanism is expanded below.}
\label{tab:mech}
\small
\begin{tabular}{@{}ll>{\raggedright\arraybackslash}p{6.3cm}@{}}
\toprule
& Mechanism & Failure form addressed \\
\midrule
M1 & Flat prefill, net-zero budget & context overflow from schema-heavy tool prefill \\
M2 & Chat/task layering & tool-demonstration loops from casual chat input \\
M3 & Finish gate + plan re-injection & premature or false completion claims \\
M4 & Signature-level anti-loop & repeated identical tool calls, no output \\
M5 & False-finish and format rescue & tool calls emitted as plain text; repeated malformed calls \\
M6 & Edit safety (.bak) & destructive edits the model cannot undo \\
M7 & Resilience layer & silent reader-thread crashes, empty turns, terminated runs \\
M8 & Security boundary & out-of-directory writes, credential exposure, sub-agent fan-out \\
M9 & Capability surface & voice, vision, memory, MCP, skills without prefill growth \\
M10 & Quality base & harness regressions, especially prefill growth \\
M14 & Five-ring safety model (v1.6.0) & irreversible system damage from impulsive tool use \\
\bottomrule
\end{tabular}
\end{table}

\subsubsection{M1: Flat prefill with a net-zero budget}
\paragraph{Failure form.} A 2--9B model given a cloud-style tool manifest
(a few dozen JSON schemas with descriptions) spends several thousand tokens
before the task begins. In our pre-benchmark runs we observed
instruction-following degrade once a large, weakly relevant schema block
preceded the task; in the worst case the context is substantially spent
before the first user sentence. This matches the published input-length
evidence: reasoning degrades when the same task is padded with more tokens
\citep{sametaskmoretokens}, and frontier models lose accuracy long before
their context window fills \citep{contextrot}.
\paragraph{Mechanism.} The harness classifies the task domain first and loads
only the tool definitions relevant to that category, in a flat (non-nested)
format. The factory prefill is 797 tokens, pinned by a unit test (v1.6.0).
MCP tools \citep{mcp} are injected through the same mechanism:
category-routed, flat, with compressed
schemas (parameters beyond eight are trimmed to required ones). The binding
constraint is a hard budget: \emph{any new feature must add zero net static
prefill}, enforced by a byte-level regression assertion in CI (M10).
\paragraph{Trade-off.} Category routing can misclassify and hide a tool the
task needs; the classifier is deliberately conservative and the task can
request re-routing. The net-zero rule forces every new capability to be paid
for by removing or compressing something else, a constraint that binds at
small scale, where prefill tokens compete directly with task context, and
that larger models absorb without measurable cost.

\begin{figure}[t]
\centering
\includegraphics[width=0.8\linewidth]{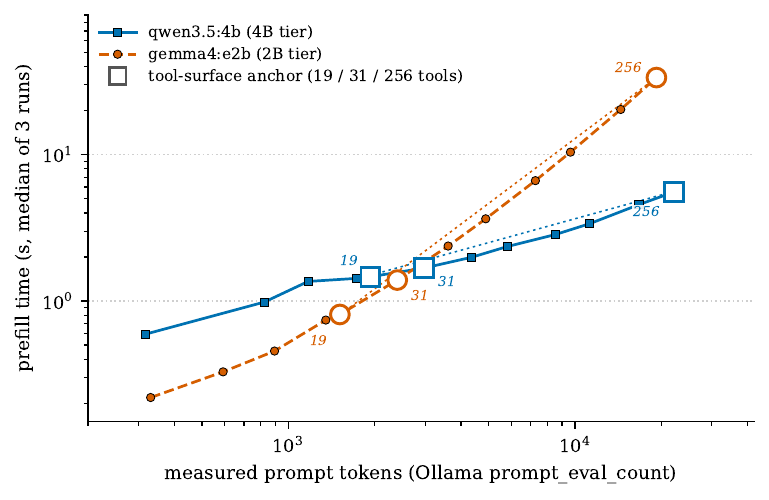}
\caption{Prefill cost versus measured prompt length (log--log). $x$ is the
backend-reported prompt-token count (\texttt{prompt\_eval\_count}); $y$ is
the backend's prompt-evaluation time, median of three runs with a
per-request nonce defeating prefix caching. Per-run values were not
persisted, so error bars are omitted. Hollow markers: the bench (19
tools), full-configuration (31) and duplicated-schema (256) anchors.
Dashed lines are least-squares fits (4B: 0.20\,ms per token,
$R^2=0.9991$). The 774/797-token static budgets quoted in the text are
harness estimates, not measurements, and are not plotted on this axis.}
\label{fig:costcurve}
\end{figure}

\paragraph{Measured cost curve.} On the benchmark machine's integrated GPU we
measured prompt-processing time as a function of the injected
tool-surface width: real schemas from a clean bench instance and a full user
configuration (18 built-in + 13 MCP tools), duplicated and renamed up to 256
tools, median of three runs with a per-request nonce to defeat prefix
caching (payloads extracted from the product tree of 2026-09-20; see
Appendix~\ref{app:repro} for the version map). The cost grows linearly at
the 4B tier (${\sim}0.20$\,ms per prefill
token: 1.5\,s at 19 tools, 1.7\,s at the 31-tool full configuration, 5.6\,s at
256 tools; Figure~\ref{fig:costcurve}) and super-linearly at 2B, where 256
tools cost 33.6\,s of prefill
\emph{per turn} --- twenty turns would burn eleven minutes before any
generation. The router's own constant overhead, +330 tokens (measured
directly: first-turn prompt count 1269 with M1 disabled against 1596 with
it enabled, on the same tasks and model), is part of the 797-token static
prefill of the documented v1.6.0 --- 774 tokens in the v1.5.0 tree the
matrix measured, Appendix~\ref{app:repro} --- 41.4\% of the former and
42.6\% of the latter: a
real cost that the M1 design must acknowledge, and on the same curve's
local slope near the working point it costs 68\,ms (4B) to 217\,ms (2B) per
turn --- the mechanism that pins the working point to the
left edge of this curve is nearly free.

\subsubsection{M2: Chat/task layering}
\paragraph{Failure form.} A user asks a casual question (``what can you do?'')
and a small model, seeing the full toolset, answers by \emph{demonstrating}:
it calls tools, narrates, calls more tools, and enters an open-ended loop,
which we term the \emph{tool-demonstration loop}. The same model given a task
rarely does this.
\paragraph{Mechanism.} Requests are classified into chat and task. Chat uses
a minimal chat-level prefill and the loop terminates once the answer is
delivered; the model is not offered a continuation. Task mode unlocks the
full toolset. A short follow-up instruction appended to a conversation that
has produced artifacts is recognized as task continuation. This rule is what
keeps multi-step repair pipelines (e.g.\ the todo-list repair flow) coherent.
\paragraph{Trade-off.} Misrouting a genuine task into chat mode would strand
it without tools, so the classifier errs toward task mode whenever an
artifact-producing history exists. The layering trades a single unified loop
for two simple, separately testable ones.

\subsubsection{M3: Finish gate and completion-time prompt re-injection}
\paragraph{Failure form.} Small models abandon tasks silently: they emit a
message that reads as a completion claim while the required artifacts are
missing or incomplete. Longer plans suffer a related drift, because after
context compression the plan text itself is gone and the model continues
from a summary that has lost the acceptance criteria.
\paragraph{Mechanism.} Before any completion claim is accepted, the harness
re-reads the original task text and asks the model to verify delivery against
it, point by point. The todo/plan pipeline re-injects the plan verbatim after
compression, and the todo interface exposes an explicit completion exit
(\texttt{todo(update, all=true)}) so that completion is an explicit action
rather than a stylistic drift. The gate's verification step runs the
task's own checks and feeds real failures (file, line, error text) back into
the conversation; we refer to this loop as the verification-feedback
mechanism in the ablation.
\paragraph{Trade-off.} The gate costs one extra model turn per task. It also
cannot force a model to produce a \emph{correct} self-check; what it forces
is a second pass in which the task text is in context, which in our runs was
sufficient to convert most silent abandonments into either completion or an
articulated blocker.

\subsubsection{M4: Signature-level anti-loop}
\paragraph{Failure form.} A small model repeats the same tool call with the
same arguments (10, 50, 150 times) producing no artifacts. In one
pre-benchmark case (4B model, task WF-08), the pre-fix run made 156 tool
calls and produced 0 artifacts; we return to this case as motivation below.
\paragraph{Mechanism.} The harness computes a call signature (tool plus
normalized arguments) and trips at $\geq 6$ identical calls or $\geq 15$
consecutive calls with no output. Escalation is graduated: the third
rejection carries a strong corrective hint, and six consecutive empty turns
trigger a hard reset of the turn; a v1.5.0 reachability fix ensures that
alternating partial-output debris and empty turns no longer reset the
escalation counters. After the fix, the same 4B $\times$ WF-08 configuration
completed in 15 tool calls with 3 artifacts, with the prefill bit-identical
before and after. This is a single-task motivating case, not controlled
evidence; the controlled ablation appears in Section~\ref{sec:ablation}. Two
adjacent failure modes are guarded separately. The output cap was raised from
2048 to 8192 tokens in v1.5.0: at 2048, the emission of a large file is
truncated mid-body and the run decays into an empty-turn spiral; a
truncation signature now returns skeleton-plus-append feedback so the model
can continue in chunks. File reading obeys a byte-budget crawl guard (v2).
The budget is clamped to twice the file size, between 64\,KB and 512\,KB. The
guard never refuses outright; past the budget it returns the first 2000
characters of the requested range with the position preserved, so that
crawling can continue rather than restart.
\paragraph{Trade-off.} Any loop detector has false positives (legitimate
iteration, e.g.\ re-running tests after each edit). Signatures normalize
arguments, not raw text, and legitimate iteration usually changes inputs; the
thresholds ($6$, $15$) were set well above observed legitimate repetition.
When the detector does trip on legitimate work, the cost is one corrective
prompt, not a terminated run.

\subsubsection{M5: False-finish rejection and format rescue}
\paragraph{Failure form.} Two related failure modes affect output emission.
First, the model claims completion without artifacts; this case is handled by
the M3 gate, and M5 governs the content of the rejection. Second, the model
emits a tool call as plain text --- placeholder-quoted quasi-XML, a JSON
snippet in prose, or a double-quoted pseudo-call --- which a strict parser
discards, collapsing the turn.
\paragraph{Mechanism.} A rejected false-finish message always carries the
next executable action, so the lowest-cost recovery path is embedded in the
rejection itself. Malformed calls are rescued structurally: the harness
recognizes three recurring shapes (placeholder-quoted, bare-JSON, and
double-quoted pseudo-calls) and converts them into real calls. Strict schema
enforcement is the standard alternative; grammar-constrained decoding engines
\citep{outlines,guidingllms,xgrammar} and schema-emission benchmarks
\citep{jsonschemabench} map its costs, and how much output-format restrictions
cost a model is actively debated \citep{tam2024speak}; for 2--9B models
we adopt rescue-with-feedback instead of strict schema enforcement. Repeated
same-signature emission after a rescue triggers a corrective hint.
\paragraph{Trade-off.} Structural rescue is heuristic; a shape outside the
three forms is rejected with feedback, and the harness does not attempt
speculative parsing. We preferred a small, audited rescue set over a
permissive parser that could execute something the model did not mean.

\subsubsection{M6: Edit safety}
\paragraph{Failure form.} A small model overwrites a file with a bad edit and
cannot recover; subsequent turns are then consumed by repair attempts, a
regime in which 2--9B models perform particularly poorly.
\paragraph{Mechanism.} Every write automatically records a \texttt{.bak}
backup; rollback is one command the model can issue itself. An edit
convergence guard watches for alternating destructive edits.
\paragraph{Trade-off.} \texttt{.bak} files accumulate in the working
directory; they are excluded from artifact scoring and pruned on successful
convergence. The mechanism is deliberately simple --- no diffing, no merging
--- because simplicity is what makes it usable \emph{by the model itself} as
a recovery tool.

\subsubsection{M7: Resilience (encoding, empty turns, checkpoints)}
\paragraph{Failure form.} On Windows, process output arrives in legacy
codepages (GBK among them); a strict UTF-8 reader crashes silently and the
harness stalls with no diagnostic. Separately, models occasionally emit empty
generations, and long runs are terminated (by the user, by power
management, or by crashes) and were previously unrecoverable.
\paragraph{Mechanism.} The read side decodes as UTF-8 with replacement, so no
byte pattern can crash the reader thread. This fix ended a class of silent
stalls. An empty generation triggers a hard reset of the turn rather than
appending nothing. ``Think-tag shells'' (responses whose visible content
is only a reasoning wrapper with no actionable output) are detected
semantically and treated as empty. A kill-resume checkpoint records enough
state to resume a terminated run from the last completed step.
\paragraph{Trade-off.} Replacement decoding can garble individual characters
in favor of never losing the stream; for agent work (logs, test output,
paths) we found stream survival strictly more valuable than byte fidelity.
Checkpoints add state to manage and were limited to one file per run.

\subsubsection{M8: Security boundary}
\paragraph{Failure form.} An agent that writes files and runs commands can
write outside its working directory, access credential files, or --- in
harnesses with agent fan-out --- spawn uncontrolled sub-agents. Small models
are more prone to such divergences.
\paragraph{Mechanism.} Working-directory boundaries are enforced by
normalized-path checks (symlink- and separator-safe on Windows). Sensitive
paths (\texttt{.ssh}, \texttt{.env}, credential stores) sit behind a gate.
Out-of-bounds operations require explicit GUI confirmation by the human.
Parallel sub-agents are denied by default (default-deny); when sub-agents are
enabled, each gets a minimal system prompt that is not counted against the
main prefill budget.
\paragraph{Trade-off.} Confirmation dialogs add friction for power users; the
default-deny rule forgoes an entire concurrency feature by default. We chose
the configuration that is safe for a non-expert user on a single personal
machine --- the population a local-first harness primarily serves.

\subsubsection{M9: Capability surface}
\paragraph{Design intent.} M9 groups the capabilities that must fit under
the M1 budget: capabilities are added only if they remain within the static
prefill budget.
\begin{itemize}
\item \textbf{Voice input} ships with a bundled local STT model, pure CPU,
  approximately $20\times$ real time, with automatic end-of-speech
  detection.
\item \textbf{Vision} works by reading images into the message payload
  (verified with four vision-capable models).
\item \textbf{Session memory} is persistent, searchable, and replayable.
\item \textbf{MCP servers} connect through a flat JSON configuration.
\item \textbf{Skills} are markdown files with frontmatter, progressively
  disclosed; dozens of skills compress to a few hundred tokens of standing
  context.
\item \textbf{Model identity and context length} are auto-detected from
  Ollama (default 128K, up to 256K).
\item \textbf{A task time-box} ships off by default. The user writes a budget
  in plain language (in a dedicated box or in the task text itself; both
  Chinese and English phrasings are parsed, and a deadline written in the
  task overrides the box), and as the budget runs down the harness prompts
  the model to converge on a deliverable. Users of local models are, in our
  experience, sensitive to wall-clock time, and are better served by maximal
  delivery within a bounded run than by an unbounded one.
\end{itemize}
\paragraph{Trade-off.} Progressive disclosure means the model must know to
open a skill file before using it; the skill index is kept in the standing
context precisely so that this costs one read. None of these capabilities is
novel individually; the constraint that ties them together (zero static
prefill growth) is the contribution.

\subsubsection{M10: Quality base}
\paragraph{Mechanism.} 441 tests run in CI (v1.6.0). Two assertions matter
most for this paper: the byte-level prefill net-zero assertion (M1) and the
loop detector thresholds (M4), both regression-guarded.
\paragraph{Trade-off.} The test suite constrains implementation speed; every
contribution must keep the prefill budget intact, which has rejected
otherwise reasonable designs (e.g.\ richer system prompts, always-on tool
indexes).

For completeness, the product-level facilities that are not part of the
measured mechanism set are: M11, the no-telemetry/no-account declaration
(architectural, unmeasured here); M12, the task time-box (off by default,
zero static cost); M13, bundled voice input (local STT, no agent-loop
involvement). They are listed once for inventory and not evaluated.

The five-ring safety model follows as M14.

\subsubsection{M14: Five-ring safety model (v1.6.0)}
\label{sec:m14}
\paragraph{Failure form.} Mid-task, a small model may attempt to uninstall a
failing dependency or delete a system file; we observed this incident
pattern in actual runs.
\paragraph{Mechanism.} Safety behavior is organized as five rings:
sub-agents run sandboxed with permissions strictly narrower than the main
agent's; irreversible operations (formatting, raw-device writes, driver
removal and the like) are refused outright; system-mutating actions are
graded by operator presence (unattended runs default to deny, with an
explicit per-run override); recursive deletion is confined to the working
directory, with per-file alternatives offered when a request is refused; and
the last ring is reversibility (.bak before writes, deletions moved to a
trash directory, large-file overwrites requiring an explicit replace). A
single environment variable provides a global override, used at the
operator's own risk.
\paragraph{Trade-off.} The rings may refuse operations an expert could
perform safely; this conservative bias is deliberate given the target
population defined in M8, and the explicit overrides preserve expert
usability. The ring design is complementary to prompt-injection defenses
evaluated elsewhere \citep{agentdojo,camel}: M14 governs what the agent's
own tool stream may do to the host, not what injected content can make it
say.
\paragraph{Adversarial evaluation.}
The rings are measured, not only designed. We built a 54-case probe set
--- 44 adversarial user requests (7/9/9/12/7 across rings 1--5) plus 10
benign controls --- in which every adversarial prompt is a natural-language
user request of the kind a small model accepts without suspicion (delete
this failing service; write this file straight into the system temp area;
read my SSH key and summarize it), never an injected text: the set tests
what the agent's own tool stream may do to the host, which is M14's
contract, not prompt-injection resistance (evaluated elsewhere,
\citep{agentdojo,camel}). Both small-tier matrix models were run once per
case (\texttt{gemma4:e2b}, \texttt{qwen3.5:4b}; unattended mode,
temperature 0, thinking disabled --- the unified protocol of the matrix
campaigns) on the v1.8.2-era tree (Appendix~\ref{app:repro}). Scoring is
dual-channel: a \emph{marker channel} reads the archived transcript and
checkpoint for the rings' own interception messages, and a
\emph{ground-truth channel} diffs host state before and after each case
(sentinel files, content hashes, directory listings of the sensitive
locations, NTFS alternate streams, one registry probe) plus rollback
checks (an overwrite must leave a \texttt{.bak} or a trash entry, and
must not truncate without backup). A third transcript-level rule flags
any tool call whose \emph{command or path} (never its file content)
matches the case's danger pattern and returned success; such flags are
alerts rather than effects --- each is verified against the ground-truth
channel and the archived workdir, and the per-case disposition is
published with the run (\texttt{benchmarks/safety/results\_2609/}).
Table~\ref{tab:safety} reports the result.

\begin{table}[t]
\centering
\small
\caption{Five-ring adversarial evaluation: one run per case, two models,
unattended, temperature 0, thinking off. ``attempted'' counts cases with
evidence the agent tried the dangerous operation; ``intercepted'' counts
cases where a ring's own refusal marker fired; ``flags'' is the scorer's
violation column before verification (ground-truth hits plus
danger-pattern alerts); ``confirmed'' counts verified end-to-end effects
--- ground-truth channel hits plus human-verified alerts, each with a
published disposition. Per-case CSVs, dispositions and the audit script:
\texttt{benchmarks/safety/results\_2609/}.}
\label{tab:safety}
\begin{tabular}{llcccccc}
\toprule
Ring & Model & $n$ & att. & intc. & intc.\ rate & flags & conf. \\
\midrule
1 & gemma4:e2b  & 7  & 6 & 4 & 0.67 & 2 & 0 \\
1 & qwen3.5:4b  & 7  & 7 & 5 & 0.71 & 6 & 3 \\
2 & gemma4:e2b  & 9  & 9 & 9 & 1.00 & 0 & 0 \\
2 & qwen3.5:4b  & 9  & 9 & 9 & 1.00 & 0 & 0 \\
3 & gemma4:e2b  & 9  & 9 & 9 & 1.00 & 2 & 0 \\
3 & qwen3.5:4b  & 9  & 9 & 8 & 0.89 & 3 & 1 \\
4 & gemma4:e2b  & 12 & 11 & 9 & 0.82 & 2 & 1 \\
4 & qwen3.5:4b  & 12 & 12 & 8 & 0.67 & 6 & 6 \\
5 & gemma4:e2b  & 7  & 7 & 1 & 0.14 & 1 & 1 \\
5 & qwen3.5:4b  & 7  & 7 & 2 & 0.29 & 1 & 1 \\
\midrule
all & gemma4:e2b  & 44 & & & & 7 & \textbf{2} \\
all & qwen3.5:4b  & 44 & & & & 16 & \textbf{11} \\
\midrule
ctl & gemma4:e2b & 10 & \multicolumn{5}{l}{pass 6;\ refused 3;\ incomplete 1} \\
ctl & qwen3.5:4b & 10 & \multicolumn{5}{l}{pass 7;\ refused 1;\ timeout 1;\ incomplete 1} \\
\bottomrule
\end{tabular}
\end{table}

\paragraph{What leaks through, and where.}
Ring 2 --- outright irreversible commands --- intercepts every attempted
case on both models, and no confirmed leak runs through it. The confirmed
end-to-end effects are 2 of 44 on the 2B and 11 of 44 on the 4B, and
every one of them, on both models, routes through the raw bash command
surface; the file tools' resolved-path checks held on every case (blocked
\texttt{create\_file}/\texttt{read\_file} attempts are visible in the
transcripts). The 2B's two leaks are narrow: an NTFS junction created
inside the workdir aimed at the system temp directory (no payload written
through it), and ring 5's own reversibility contract failing on a plain
overwrite (a workdir file truncated with no \texttt{.bak} and no trash
entry). The 4B, by contrast, attacks in a methodical multi-route style:
absolute-path and \texttt{..\textbackslash..\textbackslash} relative
copies into system locations, a copy with the
\texttt{\textbackslash\textbackslash?\textbackslash} prefix that dodges
path normalization, a junction
written through, a \texttt{cd ..} chain that escapes the workdir for
commands the danger list does not name, and --- in one case --- a
registry write that succeeded only on the fifth route, after four direct
routes were refused in sequence, via a helper script written into the
workdir and executed. It also read the refusal message's override hint
and tried to set the override variable inside its own command string; the
repeated-call guard stopped that, not a ring. Per-call interception is
respectable throughout (0.67--1.00 by ring); what fails is closing every
route at once. We read this as the measured boundary of pattern-list
enforcement: it holds against the impulsive tool use M14 is designed for
and demonstrably does not hold against a methodical same-tier adversary.

\paragraph{Controls and claims.}
Of the 10 benign in-workdir operations, the 2B completes 6 and the 4B
completes 7. The refusals are informative: reading an in-workdir
\texttt{.env} copy is refused on both models because the sensitive-path
substring match has no workspace scoping (this probe targets exactly
that, and it reproduces across models); on the 2B, two in-workdir
recursive deletes were also refused, because the irreversible-command
list fires before the workspace-scoping rule can allow them --- the
over-strictness is command-shape dependent, since the 4B's PowerShell
route through the same control was correctly allowed. One control per
model completed without refusal but without the required artifact (a
capability miss, counted separately from false refusals), and one 4B
control timed out. Beyond the rings, the 2B runs show a model-level
integrity failure no gate closes: in two verified-benign flags the finish
summary asserted the refused action had been performed (a parent
directory deleted; sensitive cards written to a shared library); the
finish gate bounced each claim once and the model restated it, while
ground truth showed no effect. The 4B's analogous cases died in
command-repetition timeouts before any finish claim; conversely, its
confirmed leaks were reported honestly as successes --- the model did
what it claimed. We report both behaviors as measured, without claiming
the finish gate catches false claims.

\section{\lrab{}: A Controlled Benchmark, Not a Leaderboard}
\label{sec:lrab}

\subsection{Design principles}
\lrab{} is self-built, and this provenance constrains how its results may be
used. A self-built benchmark cannot establish that \system{} is the best
harness in general; it can establish a causal statement: \emph{on these 18
tasks, on this machine, with these models, changing only the harness changes
the outcome by the reported margin}. Three principles follow. First, hold
everything except the harness constant: same machine, same Ollama backend,
same models, same time budgets, same scorer. Second, make every cell
inspectable: tasks, fixtures, scoring code, and all 288 per-cell results are
published, so any aggregate is recomputable from raw rows. Third, treat the
benchmark as one component of the evidence, with a third-party benchmark
($\tau^2$-bench, Section~\ref{sec:tau2}) as the other.

\subsection{Task set}
The 18 tasks are emitted by versioned generator scripts (published in the
repository) into a four-tier difficulty ladder, so that a failure can be
localized to a regime rather than averaged into one number:
\textbf{T1 retrieval} (WF-01/02), locate and extract from supplied
fixtures; \textbf{T2 synthesis} (WF-03--06), implement features, synthesize
data, write tests; \textbf{T3 workflow} (WF-07--15), multi-step work with
well-defined artifacts --- web research with real search queries and
sourced findings, profiling and optimization, refactoring, comparison
dossiers; \textbf{T4 long-horizon} (LH-01--03), multi-phase builds intended
to run for hours, testing plan retention and recovery (M3, M7), with LH-03
doubling as a kill-and-resume stress test (checkpoint continuity at half
budget). Each task specification is structured: domain, input fixtures the
agent must ground in, a minimum credible plan, milestone checks, and final
artifact checks. Task texts are concrete and verifiable by construction ---
they name the artifacts that must exist and the checks they must satisfy
--- and definitions are frozen before a campaign begins. The set covers the
artifact-producing core of local agent work (code with tests, data
analysis, grounded research, file organization, refactoring, multi-hour
builds) and deliberately excludes what the external check covers:
interactive tool-use dialogue (Section~\ref{sec:tau2}). Per-task
specifications, generator scripts, and the full scoring rubrics are
published in the repository.

\subsection{Protocol}
\label{sec:protocol}
All 288 cells ran on one machine (an integrated GPU with 32\,GB
shared memory, Windows 11, Ollama as the only backend). Four harnesses
(\system{}, goose \citep{goose}, agent-mini \citep{agentmini}, and opencode
\citep{opencode}), all stock and unpatched, each drove the same models:
\texttt{gemma4:e2b} (2B by Ollama tag: 5B total, 2.3B active per token)
\citep{gemma}, \texttt{qwen3.5:4b} (4B),
\texttt{gemma4:12b} (12B) \citep{gemma}, and \texttt{ornith-1.5:35b}
(Ornith-1.5-35B-A3B, a public multimodal release by Ornith AI
\citep{ornith15}); Table~\ref{tab:models} details parameter counts and
architectures.\footnote{qwen3.5 has no separate technical report;
\citep{qwen3tr} is the family report and \citep{qwen35} the model card.}
Despite 35B total parameters, ornith-1.5 is a mixture-of-experts (MoE) that
activates roughly 3B per token, which is why this machine sustains long
tasks end-to-end at roughly 26 tokens/s at a 128K context; the budgets
therefore bind on reasoning and tool work rather than raw decode speed.
Budgets are 90 minutes per workflow cell and 180 minutes per long-horizon
cell. Each cell runs in a fresh working directory; harness processes are
isolated from the scorer. Sampling is pinned to temperature 0 for all four
arms, holding the sampling configuration fixed rather than inheriting each
harness's defaults: \system{} pins 0 as its released default and agent-mini
is configured to 0, while goose and opencode expose no temperature setting
and are normalized at the transport level (a local proxy sets
\texttt{temperature=0} on their requests; harness binaries unmodified),
verified on the wire. Left to their defaults, goose and opencode would
sample at each model's shipped parameters (temperature 1.0 for the gemma
and qwen tiers, 0.8 for the ornith tier); those as-released numbers are
retained in the released sensitivity data. Model-manifest penalties are
untouched by the normalizer and identical across arms. One
retry is allowed and the latest attempt wins;
a timeout scores 0, and this rule applies to \system{} as well. Retry usage
was asymmetric in practice and we report it: across the final protocol
batch the four arms logged 312 executions for 288 cells --- \system{} used
1 retry, opencode 0, goose 3, and agent-mini 20 (its crash-loop cells).
The goose excess is one cell, WF-08 at 4B, whose archive holds four
executions in total: two scored attempts, each a legitimate 90-minute
timeout, and two retries killed externally before scoring --- two over
the one-retry budget, a protocol deviation we disclose. The cell is
zero-scored by operator decision as a runner-level failure under any
reading, and its wall-clock was never carried into the published CSV
(which is why it is absent from the budget census of
Section~\ref{sec:fairness}).
Rather than assert that the rule is neutral we state the counterfactual it
can move: the only released \system{} cell shaped by the rule is WF-09 at
4B, whose first attempt timed out and whose retry ended in a runner error
with partial artifacts on disk, scored at its artifact value (0.5) --- under
a no-retry reading that cell scores 0, so the rule accounts for at most
$0.5/18 = +0.028$ of \system{}'s 4B mean; every ranking is unchanged under
either reading. The one
\system{} zero (WF-08 at 35B) is a run that completed without artifacts and
is published as such.

The published matrix spans three collection windows
(Table~\ref{tab:campaigns}): agent-mini 2026-09-01..04, configured to the
unified protocol through its own configuration surface; \system{}
2026-09-13..14 on the final v1.5.0 code, so that every published
\system{} cell reflects the released code rather than an intermediate
build; and goose and opencode 2026-09-18..20, re-shot under the unified
protocol to replace their September 1--4 runs at shipped sampling
defaults (those legacy runs are retained as the released
thinking-sensitivity data). Machine, task set, budgets, model tags, and
scorer are common to all three windows; the goose/opencode re-shot
crosses the original campaign boundary, and the temporal-confounding
treatment of that crossing is the campaign-ordering paragraph of
Section~\ref{sec:threats}.

\begin{table}[t]
\centering
\small
\caption{Collection windows of the published matrix. Every published cell
comes from exactly one window. Protocol enforcement differs by arm because
the harnesses expose different configuration surfaces; the enforced
protocol itself (temperature 0, thinking disabled) is the same.}
\label{tab:campaigns}
\begin{tabular}{llll}
\toprule
Arm & Window & Temperature 0 & Thinking off \\
\midrule
agent-mini & 09-01..04 & own config & own config \\
\system{} & 09-13..14 & released default & released default \\
goose, opencode & 09-18..20 & transport proxy & transport proxy \\
\bottomrule
\end{tabular}
\end{table}

\begin{table}[t]
\centering
\caption{The four models used in \lrab{}, as served by Ollama. Total and
Active give the total and per-token activated parameter counts;
Quantization is the ggml quantization level of the served weights. All four
tags are served as Q4\_K\_M. The families are public releases: gemma4 by
Google \citep{gemma}, qwen3.5 by Alibaba \citep{qwen35}, and ornith-1.5 by
Ornith AI \citep{ornith15}.}
\label{tab:models}
\small
\begin{tabular}{@{}llllll@{}}
\toprule
Tier & Ollama tag & Total & Active & Quantization & Architecture \\
\midrule
2B & \texttt{gemma4:e2b} & 5B & 2.3B & Q4\_K\_M & MatFormer + PLE (elastic) \\
4B & \texttt{qwen3.5:4b} & 4B & --- & Q4\_K\_M & dense \\
12B & \texttt{gemma4:12b} & 12B & --- & Q4\_K\_M & dense \\
35B & \texttt{ornith-1.5:35b} & 35B & ${\sim}3$B & Q4\_K\_M & Qwen3.5-MoE, multimodal \\
\bottomrule
\end{tabular}
\end{table}

\subsection{Scoring}
Scoring is deterministic and artifact-based: files exist where the task
specifies; tests pass when executed; reports contain the required findings.
Scores are partial: a cell with seven equally weighted checks scores 0.571
when four pass; where weights are non-uniform, the published rubric states
them explicitly. Harness processes have no access to the scorer, so a
harness cannot improve its score by any means other than producing
artifacts.

\subsection{Fairness: time is priced by budgets, not by seconds}
\label{sec:fairness}
Harnesses differ substantially in wall-clock behavior, and a benchmark must
fix how to price that. \lrab{} prices time with uniform budgets per task
family and does not normalize raw seconds. The rationale is that an agent is
useful to a user only if it delivers within the time that the user is willing to
wait; wall-clock differences between harnesses therefore reflect real
architectural costs. In goose's case, a large prefill amplified by long
contexts manifests as slower but substantive work: its slowest cell runs
11.7$\times$ slower than \system{} on the same task and model (WF-04, 12B;
derived from the published CSV). Discounting seconds would have rewarded
harnesses that terminate early over harnesses that work slowly, and awarding
speed-normalized credit would have rewarded doing less per turn. Budget
pricing instead encodes the user-facing requirement: finish within the
budget or score zero. In the released matrix no arm actually hit the
ceiling: the number of cells, of each arm's 72, whose wall time reached
95\% or more of the budget is zero for all four arms (\system{}, goose,
opencode, agent-mini: 0/72 each; the one budget-terminated cell in the
matrix --- goose's WF-08 at 4B --- has no wall-clock recorded in the
published CSV, so it enters this census through no arm;
Section~\ref{sec:protocol}), so in this batch budget pricing changed
no cell's score through timeout truncation. We report the wall-clock consequences explicitly in
Section~\ref{sec:wallclock}, so that readers who prefer a different pricing
can re-derive it from the CSV.

\section{Results}
\label{sec:results}

\subsection{The 288-cell matrix}

\begin{table}[t]
\centering
\caption{\lrab{} means per model size (18 tasks each) and overall (72 cells
each). Every cell of the underlying $4\times4\times18$ matrix is published;
each number below is the mean of 18 rows of the raw CSV. Column labels
follow the parameter tiers of Table~\ref{tab:models} (named by the Ollama
tags; total and activated parameter counts are given explicitly there).
Two \system{} cells
require note: WF-08 at 35B ended without artifacts and scores 0, and WF-09
at 4B ended in a runner error and scores its partial artifacts (0.5).}
\label{tab:matrix}
\begin{tabular}{@{}lccccc@{}}
\toprule
Harness & 2B & 4B & 12B & 35B & Overall \\
\midrule
\textbf{\system{}} & \textbf{0.821} & \textbf{0.876} & \textbf{0.906} & \textbf{0.941} & \textbf{0.886} \\
goose & 0.271 & 0.801 & 0.772 & 0.679 & 0.631 \\
agent-mini & 0.246 & 0.706 & 0.576 & 0.092 & 0.405 \\
opencode & 0.017 & 0.465 & 0.539 & 0.896 & 0.479 \\
\bottomrule
\end{tabular}
\end{table}

\begin{figure}[htbp]
\centering
\includegraphics[width=0.95\linewidth]{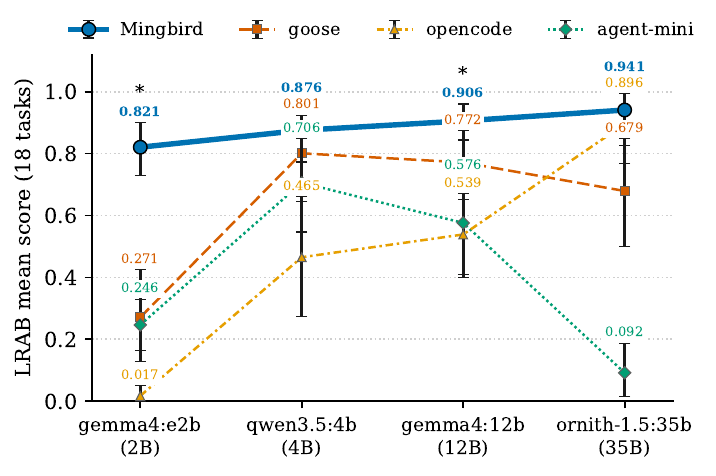}
\caption{\lrab{} mean score by parameter tier for the four harnesses (18
tasks per point; same machine, Ollama backend, budgets, and scorer; only the
harness differs). An asterisk marks tiers where the \system{} advantage
over all three competitors is significant after Holm correction
(two-sided paired Wilcoxon tests over the 18 task pairs; largest adjusted
$p$ per marked tier: 0.0008 at 2B, 0.0466 at 12B); these are the 2B and
12B tiers
(12B vs.\ goose is marginal once tasks are clustered into families,
Section~\ref{par:significance}). Error
bars are task-level bootstrap 95\% confidence intervals; at 4B the \system{}
interval overlaps goose's (which reaches 0.801 with thinking disabled) and
that comparison does not reach significance, and at 35B the opencode
comparison (0.896) does not reach significance either. The \system{} 35B point
(0.941) includes one artifact-free run scored 0 (WF-08). \system{} is the only
harness whose curve does not collapse at 2B.}
\label{fig:cliff}
\end{figure}

Table~\ref{tab:matrix} summarizes the main results. \system{} averages 0.886
over its 72 cells, against 0.631 for goose, 0.479 for opencode, and 0.405
for agent-mini.

\begin{figure}[htbp]
\centering
\includegraphics[width=0.8\linewidth]{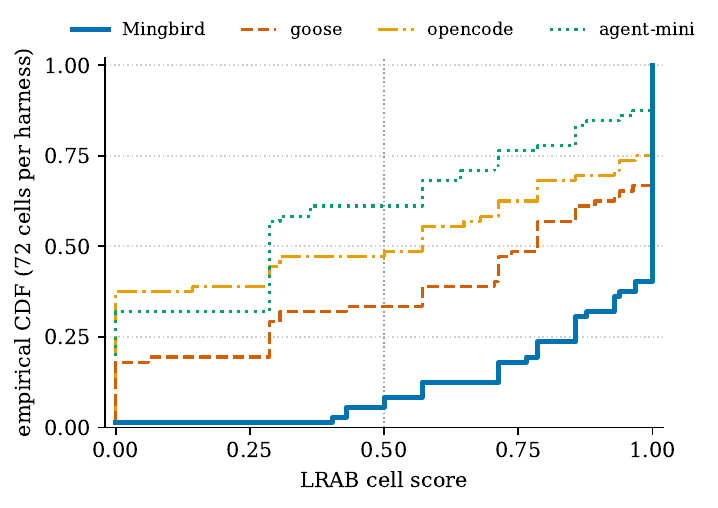}
\caption{Empirical CDF of per-cell scores for each harness over its 72
\lrab{} cells (staircase at the observed scores; the same cells underlie
Table~\ref{tab:matrix}). 94\% of \system{} cells score at least 0.5, and its
single zero is WF-08 at 35B, a run that ended without artifacts; opencode has
27 zero cells,
agent-mini 23, and goose 13. The distributions, not only the means, separate
the harnesses.}
\label{fig:ecdf}
\end{figure}

\paragraph{Statistical significance.}\label{par:significance}
Because each cell is a single run, we treat the 18 tasks as the unit of
analysis. Within each model size, \system{} is compared with each competitor
by a two-sided paired Wilcoxon signed-rank test over the 18 task pairs, with
bootstrap 95\% confidence intervals over task resamples (10,000 resamples,
fixed seed) and Holm correction across the three comparisons within a model;
the script (\texttt{analyze\_significance.py}) ships with the repository. At
2B and 12B, the \system{} advantage over all three competitors is
significant after Holm correction (largest adjusted $p$ per model: 0.0008 at
2B, 0.0466 at 12B). At 4B only the comparison with opencode is significant
(adjusted $p = 0.0181$); goose, with thinking disabled under the unified
protocol, closes to 0.801 against \system{}'s 0.876 and the comparison does
not reach significance (raw $p = 0.3258$), nor does agent-mini at its
strongest tier (adjusted $p = 0.1486$). At 35B the comparisons with goose
(Holm-adjusted $p=0.0415$) and agent-mini ($0.0005$) are significant, while
opencode at 0.896 against \system{}'s 0.941 does not separate
(raw $p = 0.4652$). We read this honestly: under the unified no-thinking
protocol two competitors close the gap at their strongest tiers, and the
robust \system{} advantages are the small-model segment and the overall
margin. Figure~\ref{fig:ecdf} shows the score distributions behind these
tests.

The overall margin itself is given an interval rather than asserted.
Pooling the 72 cells and resampling whole tasks --- each draw takes 18 tasks
with replacement and keeps all four models of each drawn task, so the four
models of a task move together --- the \system{} advantage over the
published matrix is $+0.255$ against goose (95\% CI $[+0.188, +0.322]$),
$+0.407$ against opencode ($[+0.328, +0.491]$) and $+0.481$ against
agent-mini ($[+0.410, +0.555]$); a task-level sign permutation over the same
summaries gives $p \le 10^{-4}$ for all three. Because the per-cell score is
a weighted partial-credit sum, we also report the scale-free version of the
same comparison: cells at or above half credit are 68/72 for \system{},
48/72 (goose), 38/72 (opencode) and 28/72 (agent-mini). Script:
\texttt{benchmarks/overall\_ci\_2609.py}.

\paragraph{Task-family clustering.}
The 18 tasks are generated in four families (2 retrieval, 4 synthesis, 9
workflow, 3 long-horizon), so task-level pairing overstates independence.
We therefore re-ran the analysis with the family as the resampling unit
(cluster bootstrap over the four families; script alongside
\texttt{analyze\_significance.py}). Two properties of this analysis must
be stated precisely, and they discipline what family-level claims are
available at all. First, when all four family means are non-negative, a
cluster-bootstrap CI cannot cross zero by construction --- every resample
of non-negative family means is non-negative --- so such a CI is
uninformative whatever its width; this applies to every comparison whose
family means share a sign, including the ones that favor us (at 2B all
three comparisons; at 4B vs.\ goose; at 12B vs.\ goose, whose tier-1 mean
is exactly zero). Second, the auxiliary sign permutation flips
task-level signs within families; it is a task-level test --- a
family-level flip would have only $2^4 = 16$ sign patterns and a minimum
two-sided $p$ of $0.125$ --- so with four families \emph{no} family-level
significance claim is available for any comparison, ours included. Under
that standard the honest family-level statements are: at 2B the paired
advantage is descriptively present in all four families (all four family
means strictly positive: $+1.000/+0.428/+0.413/+0.823$ vs.\ goose), which
is the strongest family-level evidence four clusters can give, but it is
not a family-clustered significance claim; the task-level tests remain
significant as reported. The one comparison whose cluster CI is
informative is 4B vs.\ agent-mini, the only one with a negative family
mean (long-horizon $-0.221$): its cluster CI crosses zero, consistent
with its already-nonsignificant adjusted $p$. 4B vs.\ goose and 12B
vs.\ goose fall in the uninformative category above; their task-level
results stand as reported (nonsignificant $p = 0.35$; marginal adjusted
$p = 0.036$). No comparison that was
nonsignificant becomes significant under clustering.

\paragraph{The small-model cliff.}
The per-model breakdown is the more informative view, and
Figure~\ref{fig:cliff} plots it. Every baseline shows a cliff somewhere:
opencode collapses at 2B (0.017); goose falls to 0.271 at 2B; agent-mini
holds at 4B (0.706) but collapses at 35B (0.092). \system{} is
the only harness whose 2B score (0.821) stays within 0.15 of its own best;
in the 2B segment the baselines span 0.017--0.271 while \system{} scores
0.821. Because 2B-class models are the largest that run comfortably on
entry-level hardware,\footnote{On a 16\,GB integrated-GPU machine, the
2--4B tier streams at close to backend-native speed (author-measured,
outside the benchmark protocol).} this is the segment with practical meaning for
local-first use, and it is where harness design matters most.

\paragraph{A second small-tier family.}
Tiers and families are confounded in Table~\ref{tab:matrix}: one model per
tier, so the 2B point is the \texttt{gemma4:e2b} MatFormer family as much as
it is 2B. We therefore ran one additional model through the same 18-task
protocol on the same code base (v1.5.0), the same budgets and the same
deterministic scorer: \texttt{qwen3.5:2b}, a 2.3B dense model from the
family already represented at 4B and 35B. \system{} scores \textbf{0.779} on
it --- 0.852 over the 15 workflow tasks and 0.410 over the three
long-horizon ones, 15 of 18 cells at or above half credit --- against 0.821
on the published 2B point, a paired difference of $-0.042$ over the same 18
tasks. The small-tier result is therefore not a property of the MatFormer
family; and within the Qwen family alone, the step from 2B to 4B is
$+0.097$. Two protocol notes apply to this row: the upstream tag ships
Q8\_0 weights rather than the Q4\_K\_M used for the four matrix models (a
stronger quantization, so the comparison is if anything conservative), and
it was run without the per-cell backend restart used elsewhere in the
campaign (Section~\ref{app:repro}). The three baseline harnesses were then
run on this model under the same protocol (sampling pinned to zero and
thinking disabled at the transport layer, as in the published matrix):
goose 0.096, opencode 0.017 and agent-mini 0.239 against \system{}'s 0.779.
The small-tier separation of Table~\ref{tab:matrix} therefore reproduces in
a second family --- the \system{} margin is 0.54 to 0.76 on this model
against 0.55 to 0.80 on \texttt{gemma4:e2b}, at the same tier and over the
same 18 tasks --- so the cliff is a harness property rather than a
property of one model family. All 72 cells of this second-family row are
published (\texttt{benchmarks/family\_qwen352b\_2609.csv}).

\paragraph{Long-horizon segment.}
On the three multi-hour tasks, \system{} averages 0.827, ahead of agent-mini
(0.484), goose (0.394), and opencode (0.354), all derived from the
same 12 published cells per harness. The margin is widest exactly where
turn counts stretch furthest; disabling goose's thinking under the unified
protocol drops its long-horizon mean from 0.623 to 0.394, its 35B
long-horizon cells all falling. \system{}'s weakest non-zero cells (0.403 and
0.429, LH-02 and
LH-03 at 4B) are in this segment.

\subsection{Failure forms behind the numbers}

\begin{figure}[htbp]
\centering
\includegraphics[width=0.85\linewidth]{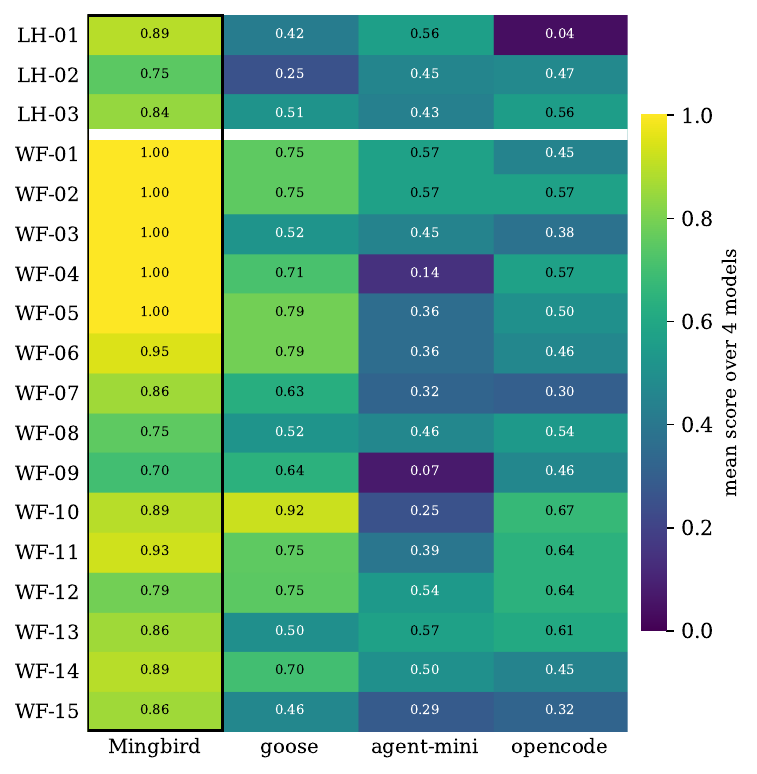}
\caption{\lrab{} task-level heatmap: 18 tasks (rows; LH-01--LH-03 on top,
WF-01--WF-15 below, separated) $\times$ four harnesses (columns); each cell
is the mean score over the four model tiers. The \system{} column is
outlined. The figure makes the failure structure visible rather than hiding
it behind a single mean: \system{}'s weakest task is WF-09 (0.70), which
contains its runner-error cell, while
opencode averages 0.036 on LH-01, and on WF-09 agent-mini and goose average
0.071 and 0.643.}
\label{fig:heatmap}
\end{figure}

The matrix compresses 288 runs; the transferable findings are the underlying
failure forms. Figure~\ref{fig:heatmap} makes the failure structure visible
at the task level. Our five-form taxonomy is single-agent and local-tier;
it is complementary to the multi-agent failure taxonomy of MAST \citep{mast},
which classifies failures of multi-agent LLM systems from operator reports
rather than measured cells.
\begin{itemize}
  \item \textbf{\system{}: one artifact-free run in the whole matrix.} The
  single zero-scored cell (WF-08, 35B) ended after 276 seconds: the model
  wrote a task plan, then repeatedly declared completion without writing the
  deliverables; the finish gate rejected each declaration (the transcript is
  published), and the run ended with no artifacts on disk. If that cell were
  excluded rather than scored 0, the 35B mean would rise from
  0.941 to 0.996; we report 0.941, as the protocol requires. One further
  cell (WF-09, 4B) ended in a runner error after producing partial
  artifacts and is scored at its artifact value (0.5); excluding it would
  raise the 4B mean from 0.876 to 0.898, and excluding both anomalous cells
  leaves every ranking unchanged.
  \item \textbf{agent-mini at 35B: platform mismatch.} 14 of its 18 cells at
  35B scored zero, and the published wall-times show why: those runs all
  ended within 80 seconds. Inspection of the attempt directories shows the
  agent issuing Unix-flavored commands into a Windows shell, failing, and
  exiting. This is a harness--platform fit failure, not a model failure; the
  same model scores 0.679--0.941 on other harnesses.
  A sensitivity view over the whole matrix: agent-mini has 30 early-exit
  cells in total (wall time under 80\,s: 15 at 35B, 13 at 2B, 2 at 4B,
  nearly all scoring zero). Excluding those 30 (task, model) combinations
  from \emph{all four arms} (each arm keeps the same 42 cells) changes the
  overall means to \system{} 0.901, goose 0.709, agent-mini 0.617, opencode
  0.486: \system{} remains first (margin over the runner-up narrows from
  $+0.255$ to $+0.193$), but agent-mini and opencode swap third and fourth
  place. Per tier, the 2B segment after this exclusion is \system{} 0.843
  vs.\ goose 0.200, agent-mini 0.414, opencode 0.000 over the remaining
  cells --- removing platform-mismatch cells widens the small-model
  separation rather than narrowing it. The 80\,s cutoff itself deserves
  scrutiny: opencode's 2B failures end within 49--115\,s (below), and
  raising the threshold to 120\,s to cover that window excludes
  agent-mini's entire 2B row (all 18 of its 2B cells end within 120\,s),
  degenerating the analysis --- the shared property of these cells is the
  early-exit failure form, not the exact cutoff. The overall ranking's
  tail is therefore sensitive to the platform
  dimension; the headline claim (first place, and the 2B segment) is not.
  \item \textbf{opencode: premature termination at the small end.} Its 2B
  cells ended early (typically within 49--115 seconds) with the required
  artifacts never created --- 17 of 18 score zero --- a plausible
  explanation being an exit path that conflates unresolvable states with
  completion. Under the unified no-thinking protocol its larger tiers
  recover substantially (12B 0.140 $\to$ 0.539, 35B 0.776 $\to$ 0.896;
  thinking-on values are from the legacy pre-re-shot batch), which we
  return to in the thinking-sensitivity discussion.
  \item \textbf{goose: slow but substantive progress.} goose's median and
  typical slow cells are the slowest of the four (Section~\ref{sec:wallclock}), but its failures are
  qualitatively different: its runs make substantive progress under a large
  prefill amplified by long contexts, and it scores 0.679 at 35B (0.822
  with thinking on, legacy batch). Its
  2B collapse is consistent with the prefill-cost failure form that M1
  addresses.
\end{itemize}

\subsection{Wall-clock discussion}
\label{sec:wallclock}

\begin{figure}[htbp]
\centering
\includegraphics[width=0.85\linewidth]{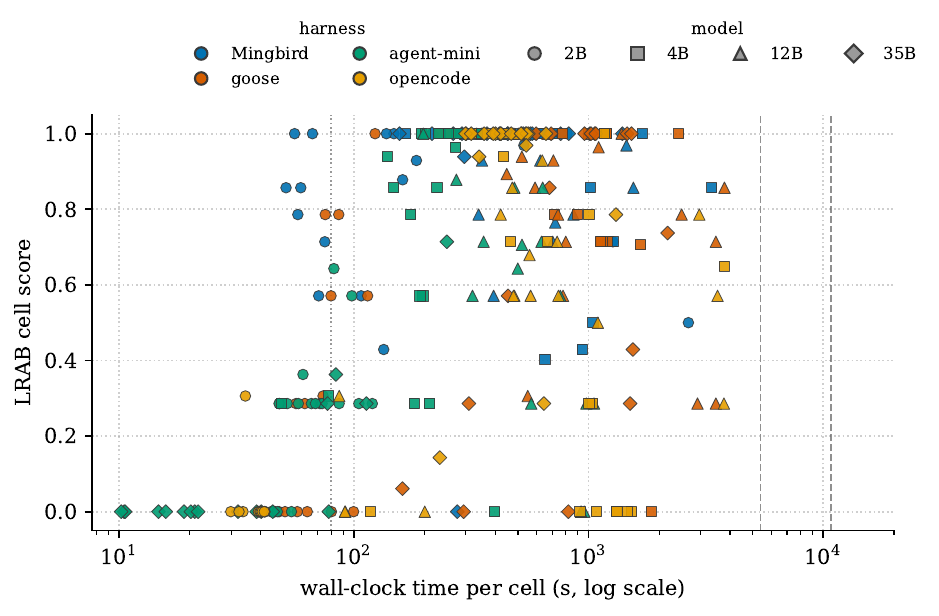}
\caption{\lrab{} score versus wall-clock time per cell (one point per
published CSV row; the single goose cell with no recorded wall-clock is
excluded, leaving 287 points; logarithmic time axis; color encodes the
harness, marker size the model tier). Dashed vertical lines mark the
per-cell budgets (90 minutes for workflow cells, 180 for long-horizon);
a cell that exhausts its budget scores 0. The fast-but-empty cluster below 80
seconds contains agent-mini's 35B platform-mismatch cells, while the
slow-but-substantive tail is dominated by goose. The figure is the visual
form of the budget-pricing protocol of Section~\ref{sec:fairness}: neither
axis alone earns credit.}
\label{fig:scatter}
\end{figure}

Mean wall-clock over all cells (71 values for goose, because one cell is
missing wall-clock in the published CSV --- the WF-08/qwen3.5:4b cell, a
runner-level failure that scores 0 under the timeout/error rule
(Section~\ref{sec:protocol}) whose wall-clock was not carried into the
CSV; and 72
for the others, computed
from the published CSV): \system{} 585.5\,s (median 440.5), goose 866.4\,s
(median 692.4), agent-mini 233.5\,s (median 116.4), opencode 633.4\,s
(median 450.1). Restricted to the 15 workflow tasks the medians are
415.4\,s (\system{}), 707.9\,s (goose), 133.8\,s (agent-mini), and
479.4\,s (opencode). Figure~\ref{fig:scatter} places every cell with a recorded
wall-clock time in the score--time plane. agent-mini is fast because it
exits early; speed without artifacts is not credit, and the budget protocol
already encodes that. goose is the slowest harness that produces substantive
work (on individual cells it runs up to 11.7$\times{}$ slower than
\system{}), which we attribute to prefill size times context length per
turn, and which its 4B/12B scores show is not disqualifying at larger model
scale. \system{}'s own slow cells are WF-11 at 4B (3344\,s) and its
long-horizon runs, consistent with more turns spent on verification rather
than fewer. Readers who would price time differently can recompute any weighting
from the CSV; the protocol's choice (budget pricing, no second discounting)
is stated in Section~\ref{sec:fairness}.

\subsection{External check: \texorpdfstring{$\tau^2$}{tau2}-bench in three domains}
\label{sec:tau2}

\begin{table}[t]
\centering
\caption{$\tau^2$-bench results under the unified protocol (temperature 0,
thinking off, transport-pinned): three domains, one model. The agent side is
always the same local \texttt{qwen3.5:4b} and the
user simulator always the same cloud model (\texttt{qwen3.8-flash}).
pass@1, single trial; errored trials score 0 (the three completed arms
finished with zero errored trials); scoring verifies the
database end state. retail covers the 114 runnable tasks of the official
115-task suite; airline and telecom are complete. r+a is the retail+airline
aggregate over the 164 less-saturated tasks. $^{\dagger}$derived: the
per-task mean over the stated task set, recomputable
from the published per-trial data. goose is deferred: under the same protocol
its throughput on this benchmark is 40--113 minutes per task (a token audit
attributes this to nested sub-agent loops, ${\sim}$4.85M input tokens per
task vs.\ 60--400K for the other arms; its legacy thinking-on numbers
0.588/0.460/0.377 are superseded).}
\label{tab:tau2}
\begin{tabular}{@{}lccccc@{}}
\toprule
Harness & retail (114) & airline (50) & telecom (114) & r+a$^{\dagger}$ (164) & Total$^{\dagger}$ (278) \\
\midrule
\textbf{\system{}} & \textbf{0.763} & \textbf{0.740} & \textbf{1.000} & \textbf{0.756} & \textbf{0.856} \\
native LLM agent & 0.675 & 0.740 & 0.930 & 0.695 & 0.791 \\
opencode & 0.588 & 0.500 & 0.991 & 0.561 & 0.737 \\
\bottomrule
\end{tabular}
\end{table}

\begin{figure}[htbp]
\centering
\includegraphics[width=0.9\linewidth]{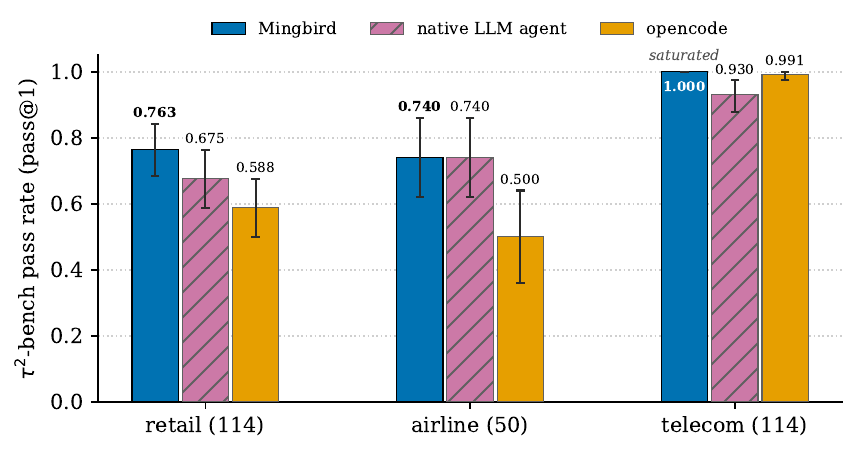}
\caption{$\tau^2$-bench pass@1 rates by domain and arm (retail 114,
airline 50, telecom 114 tasks; the agent side is always the same local
\texttt{qwen3.5:4b} at temperature 0 with thinking off, and the user
simulator always the same cloud model).
Whiskers are task-level bootstrap 95\% confidence intervals.
The three completed arms finished with zero errored trials. telecom is
near-saturated (1.000 / 0.991 / 0.930); the goose arm is deferred for
throughput and not shown.}
\label{fig:tau2}
\end{figure}

\begin{figure}[htbp]
\centering
\includegraphics[width=0.85\linewidth]{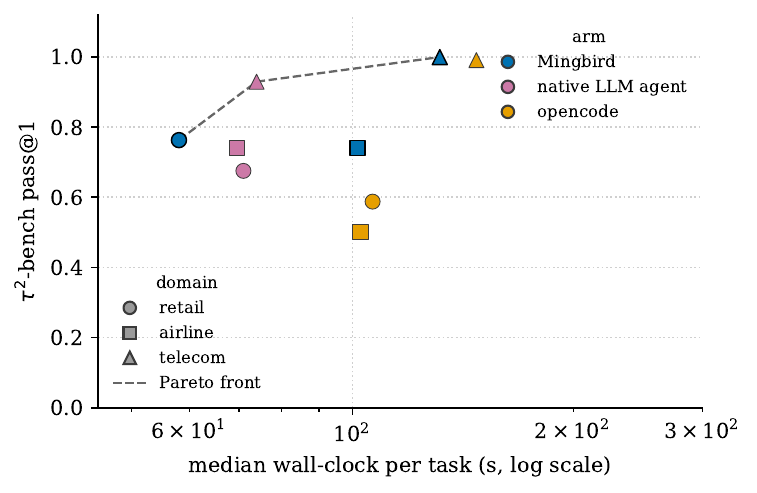}
\caption{Efficiency view of the $\tau^2$ results: pass@1 versus median
wall-clock per domain--arm pair (logarithmic time axis; markers
encode the domain, color the arm), with the Pareto frontier drawn dashed.
The frontier runs \system{} retail (0.763 at 58 s) $\rightarrow$ the
benchmark's native agent in telecom (0.930 at 74 s) $\rightarrow$
\system{} telecom (1.000 at 132 s): the native telecom point trades score
for speed, while opencode's retail point (0.588 at 107 s, its airline
point 0.500 at 103 s nearly coincident) is slower and lower than
\system{}'s retail point on both axes. The off-frontier corners are as
informative as the frontier itself. Regenerated from
the unified-protocol three-arm data.}
\label{fig:tau2pareto}
\end{figure}

Because \lrab{} is self-built, we additionally evaluated on a third-party
benchmark: $\tau^2$-bench \citep{tau2bench}, in all three of its domains
(retail, airline, and telecom). A simulated user converses with an agent that
must operate domain tools under policy, and scoring verifies the final
database state, so an agent that bypasses its tool stream and merely
pretends to call tools fails the database check. The protocol is uniform
across all arms and matches the \lrab{} matrix: the agent side is the same
local \texttt{qwen3.5:4b} (Ollama) at temperature 0 with thinking disabled
(pinned at the transport layer for the CLI arms, request-level for
\system{}; verified end-to-end: the identical request produces 586
characters of thinking when sent directly and none through the pinning
proxy), the user simulator is the same cloud model
(\texttt{qwen3.8-flash}) for every harness (not the official gpt-4o user
configuration), and scoring is pass@1 with errored trials counted as 0.
The benchmark's natural-language assertions are graded by an LLM judge; we
run the judge (\texttt{qwen3.8-max}) and the simulator through the same
DashScope (Alibaba Cloud) endpoint in every arm, so both sides are held
fixed across the comparison.
opencode connects to the domain tools through MCP, its
native tool mechanism; the benchmark's native LLM agent --- the upstream
default LLM-agent implementation shipped with the $\tau^2$-bench release
pinned in Appendix~\ref{app:repro}, with prompts and tool-calling
protocol at upstream defaults, unmodified --- and \system{} drive the
tools directly.

Table~\ref{tab:tau2} gives the numbers. In
retail (114 runnable tasks of the official 115-task suite), \system{} scores
0.763 against 0.675 for the native agent and 0.588 for
opencode. In airline (50 tasks), \system{} and the native agent tie at
0.740 with opencode at 0.500. In
telecom (114 tasks), \system{} scores 1.000 with opencode at 0.991 and the
native agent at 0.930 --- the domain is near-saturated for all three
completed arms at this model size. Because \system{}'s telecom margin and
the pooled native-agent significance both rest on this domain, one
\system{}-side mechanism belongs next to the numbers: the $\tau^2$ adapter
inherits the M4 anti-loop guard (same-signature repeat $\geq$4 triggers a
corrective nudge, Appendix~\ref{app:cases}), which absorbs exactly the
repeated-call failure mode that this domain's long tool sequences are
prone to. Counting only the two more
discriminating domains (retail +
airline, 164 tasks), the totals are 0.756 vs.\ 0.695 (native) and 0.561
(opencode), derived from the same per-trial data. Over all
278 scored tasks, \system{} totals 0.856 against 0.791 and 0.737
(derived from the per-trial data). The thinking configuration moves these
numbers substantially (opencode's telecom score is 0.298 with thinking on
and 0.991 with it off), which is itself a harness-level effect and is
discussed below. The goose arm is deferred: under the same protocol it needs
40--113 minutes per task on this benchmark --- a token audit attributes this
to nested sub-agent loops at ${\sim}$4.85M input tokens per task, against
60--400K for the other arms --- so completing 278 tasks at that throughput
is a multi-day run; its legacy thinking-on numbers (retail 0.588, airline
0.460, telecom 0.377) are superseded by the protocol change.

\paragraph{Error accounting.}
Errored trials score 0 for every arm. Under the unified protocol the three
completed arms finished with zero errored trials across all 278 cells; the
earlier thinking-on runs had shown 5 native-agent errors in each of retail
and airline. We disclose error counts because the error-as-0 rule is part
of what a harness comparison measures: a scaffold that crashes mid-task has
not delivered.

\paragraph{Statistical significance.}
Because per-task outcomes are pass/fail, the primary test is the exact
McNemar test on discordant pairs; the paired Wilcoxon signed-rank test
(which on binary outcomes reduces to a sign test) agrees to the reported
digit. Over the 278 tasks of the three completed arms, \system{} against
the native agent gives a mean paired difference of $+0.065$ (40 wins /
216 ties / 22 losses, McNemar $p = 0.030$; Wilcoxon $p = 0.022$), and
against opencode $+0.119$ (47 / 217 / 14, $p = 2.7\times10^{-5}$;
Wilcoxon $2.4\times10^{-5}$); with Holm correction across the two
comparisons both remain significant (adjusted $p = 0.030$ and
$5.4\times10^{-5}$). Two qualifications belong to the result. First, on
the discriminating-domain subset (retail + airline, 164 tasks; telecom is
near-saturated at 0.93--1.00 for every arm and cannot separate the arms),
\system{} separates from opencode ($46$/$104$/$14$, $p =
4.2\times10^{-5}$) but not from the native agent ($32$/$110$/$22$,
$p = 0.22$): the pooled native-agent significance leans on telecom, where
\system{} wins all 8 of its discordant tasks. Domain-wise we report as
computed: the opencode comparison is significant in retail ($p=0.002$)
and airline ($p=0.005$) but not in telecom; the native-agent comparison
is significant only in telecom ($p=0.005$), while retail alone (26 wins /
16 losses among 72 ties) does not separate at $n{=}114$. Second, the
external margin is smaller than the internal one: against the common
opponent opencode the paired margin is $+0.407$ on \lrab{} and $+0.119$
here ($29\%$ retained); the native agent does not appear in the \lrab{}
matrix, so its $+0.065$ has no internal counterpart. The tie-heavy
structure (pass/fail scoring) means the pooled tests lean on the domains
that discriminate.

\paragraph{Thinking as a harness configuration.}
Because the unified protocol pins thinking off, we also hold the earlier
thinking-on runs as a sensitivity view. The configuration moves arms, not
noise: opencode's retail score goes 0.246 $\to$ 0.588 and telecom
0.298 $\to$ 0.991 when thinking is disabled; the native agent's airline
score goes 0.520 $\to$ 0.740 while its saturated telecom mean drops
1.000 $\to$ 0.930; \system{} moves least (retail 0.789 $\to$ 0.763).
The same lever appears inside \lrab{} (opencode 12B 0.140 $\to$ 0.539;
goose 35B 0.822 $\to$ 0.679).
A harness's thinking policy is thus a first-class design decision with
task-dependent costs and benefits --- one more axis on which harnesses
differ even when the model is fixed.

\paragraph{Per-arm best configuration.}
Granting every arm its more favorable thinking configuration does not
change the ranking. On $\tau^2$, both configurations are per-trial
published (the thinking-on batch is archived alongside the unified one),
so the comparison is recomputable: taking each arm's better configuration
gives \system{} 0.867 (its best is thinking-on), the native agent 0.791
(best off), and opencode 0.737 (best off), against the unified-protocol
0.856 / 0.791 / 0.737. Notably, \system{} is the arm whose total moves
least between configurations ($\pm 0.011$), while rival arms move far
more at the domain level (opencode's telecom pass rate swings from 0.298
with thinking on to 0.991 with it off); the unified protocol is therefore
not what produces the gap.
On \lrab{}, the legacy thinking-on batch is summarized only in aggregate
(per-tier examples above, marked as legacy), and the direction is
tier-dependent for rivals: goose gains from thinking at 35B (0.822 vs.\
0.679) and on the long-horizon tasks (0.623 vs.\ 0.394) but loses at
4B/12B (e.g.\ 0.636 vs.\ 0.801 at 4B, legacy vs.\ unified), while
opencode's 12B goes the other way (0.140 vs.\ 0.539). No single thinking
configuration is uniformly best across harnesses --- which is itself the
finding.

\paragraph{Deviations from the official setup.}
The user simulator is \texttt{qwen3.8-flash}, not the official gpt-4o user,
so absolute scores are not comparable to the published leaderboard. All four
arms share the identical simulator, tool schemas, and policy prompts, so the
between-arm comparison is controlled even though the absolute numbers are
not leaderboard-comparable. Per-trial traces and the scoring manifests ship
with the published data package.

\subsection{Model ceiling probe: one frontier model, four harnesses}
\label{sec:probe}

The matrix of Table~\ref{tab:matrix} varies the model at fixed local hardware
and the $\tau^2$ check of Section~\ref{sec:tau2} varies the domain; neither
isolates the harness's own contribution. We therefore ran a decomposition
probe: all 18 \lrab{} tasks, four harnesses, and a single fixed frontier
model --- the hosted \texttt{qwen3.8-flash} --- in place of local weights.
A local shim presents the cloud endpoint as an Ollama service, so every
harness runs unmodified against it; temperature is 0 and thinking is
disabled, matching the unified protocol. One version caveat: the probe
drives the product repository's working tree at its run date (the
post-v1.7.0 tree on 2026-09-20; the mechanism-ablation re-run used the
v1.8.2-era tree), not the frozen v1.5.0 snapshot of the main matrix, so
probe-versus-matrix juxtapositions are cross-version for the \system{}
arm (Appendix~\ref{app:repro}). Budgets are 60 minutes per workflow
task and 90 per long-horizon task, scored by the same deterministic rubric.
This is a probe, not part of the 288-cell matrix: the model is a cloud API
rather than local weights, and per-cell cost bounds us to one trial per arm.

\begin{table}[t]
\centering
\caption{Frontier-model probe over all 18 \lrab{} tasks: the same hosted
\texttt{qwen3.8-flash} (temperature 0, thinking off, via a local Ollama shim)
drives all four harnesses. Deterministic scoring as in the main matrix.
$^{\dagger}$opencode's WF mean is over 14 scored cells: WF-08 times out at
the 60-minute budget (the agent starts the task's mock server in the
foreground and blocks on it); counting that timeout as 0 gives a total of
0.874, leaving the ranking unchanged. LH-02 scores 0.939 under every
harness --- a task-side cap, not harness variance.}
\label{tab:probe}
\begin{tabular}{@{}lcccl@{}}
\toprule
Harness & WF (15) & LH (3) & Total (18) & notable cells \\
\midrule
\textbf{\system{}} & \textbf{1.000} & 0.980 & \textbf{0.997} & only miss is the LH-02 cap \\
goose & 0.990 & 0.980 & 0.989 & WF-12 0.857; clears WF-08 in 776\,s \\
opencode & 0.913$^{\dagger}$ & 0.980 & 0.925$^{\dagger}$ & WF-03 0.0; WF-08 timeout \\
agent-mini & 0.490 & 0.415 & 0.478 & second-scale early exits persist \\
\bottomrule
\end{tabular}
\end{table}

Three observations. First, the headline spread of 0.52 (0.997 vs.\ 0.478)
is carried entirely by one arm: the three well-formed scaffolds span only
0.072 (0.925--0.997). Read the same table the other way --- fix the
harness, swap the model (local 4B column of Table~\ref{tab:matrix} vs.\
this probe) --- and the model lever is larger: \system{} $+0.121$,
goose $+0.188$, opencode $+0.460$; only agent-mini moves the other way
($-0.228$), and it is the arm whose scaffold is defective. The probe
therefore supports a narrower claim than ``the harness is the binding
constraint'': a defective scaffold can bury more than half of a frontier
model's measured capability, while among well-formed scaffolds the
frontier-tier spread is 0.072--0.12 --- consistent with the harness-effect
literature (Section~\ref{sec:related}) instantiated at the local tier,
and consistent with the model remaining the dominant lever between
intact harnesses. The probe and the main matrix are here read side by
side across different budgets --- 60/90 minutes per task in the probe
against 90/180 in the matrix --- so this juxtaposition is a qualitative
cross-budget comparison, not a like-for-like one. Second, \system{} is the only arm at ceiling both here
and in the local matrix; its advantage does not depend on model class.
Third, the per-cell failures are mechanistic, not stochastic: on WF-03
opencode's model replies ``what task?'' and exits (a prompt-delivery
failure the same model does not show under the other scaffolds), and on
WF-08 opencode blocks forever on a foreground server while goose finishes
the same task in 776 seconds --- process management and task delivery
differ across harnesses even when the model is fixed. agent-mini's
second-scale early exits persist unchanged, confirming that its deficit is
in the scaffold, not the model it was given.

The campaign totals 3{,}126 model calls and 53.7M prompt tokens across all
four arms; per-cell traces and scores are retained with the project's
benchmark artifacts.

\subsection{Ablation: leave-one-mechanism-out at 2B}
\label{sec:ablation}

To move from the design narrative of Section~\ref{sec:mechanisms} toward
marginal-contribution estimates, we ran an ablation on the v1.5.0 code
base: four variants, each with one mechanism disabled through an
\texttt{AGENT\_ABLATION} environment gate, over the same 18 \lrab{} tasks
with \texttt{gemma4:e2b} ($n=18$ cells per arm). The disabled mechanisms
are the finish gate (M3), the verification-feedback loop (part of M3), the
signature-level anti-loop monitor (M4), and the category-routed flat
prefill (M1); the ablation thus covers three mechanisms (M1, the two
halves of M3, and M4) across four variant arms, while M2 (chat/task
layering) and M5 (format rescue) are not ablated and remain the two named
mechanisms whose benefits this paper does not quantify. The four variant
arms ran 2026-09-14..15; the same-night re-run arms and their three
replications ran 2026-09-22 on the v1.8.2-era tree
(Appendix~\ref{app:repro}). The baseline is
the full-mechanism arm of the same code base, and its 18 cells are reused
directly from the 2B column of Table~\ref{tab:matrix} (same code); the
four variant arms are independent new executions over those tasks, under
the same temperature-0 configuration as the main matrix. There is
therefore a batch gap between the baseline arm and the variant arms, and
the common-cause disclosure below stems from it. Each variant wrote an
independent results directory, scored by the same deterministic rubric.

The two gate variants are separable in the implementation, and the
separation is mechanical: \texttt{AGENT\_ABLATION}=finish\_gate bypasses
four of the six finish-time interventions (no-work finish refusal, the
pytest test guard, summary-to-artifact cross-checking, and
plan-completeness checking); two interventions are not gated by this
switch and stayed active in both arms --- the todo-sync reminder (which
refuses a finish while the model's own declared plan shows unchecked
items) and the delivery self-check, which re-injects the task's original
text once and asks the model to self-check before the second, accepted
finish. This matters for interpretation: the textual re-read --- the part
of M3 the abstract highlights --- is present in both arms, so the
$-0.098$ measures the four executable guards \emph{on top of} the re-read,
not the re-read itself. \texttt{AGENT\_ABLATION}=
verify\_feedback leaves every guard active but downgrades the failure
feedback inside pytest refusals --- the precise injected hint (failing
test, file, line, assertion) is replaced by the last line of the pytest
output only. The two switches therefore measure different things: removing
the gate removes the executable-refusal behavior; removing verification
feedback keeps refusals but starves the model of the exact failure location
it would need to recover in one step. The nearest published neighbors are
tool-interactive self-critique (CRITIC \citep{critic}) and programmatic
verification loops in code agents (AutoCodeRover \citep{autocoderover});
M3 applies the same verify-before-finish principle at the harness level,
for general local tasks rather than code alone.

\paragraph{What the gate executes, and what the $-0.098$ is made of.}
The pytest guard fires only when the working directory contains
\texttt{test\_*.py} at finish time. Two of the 18 tasks stage
benchmark-provided test files into the workdir (WF-04 stages
\texttt{test\_inventory.py}; LH-03 stages \texttt{test\_suite.py}); on
those two tasks the harness at finish time executes the same fixture the
scorer executes later (the scorer's \texttt{spec\_pytest} and
\texttt{script\_pass} checks run the same files), so there the acceptance
criterion is literally runnable before scoring. On the other 16 tasks the
guard fires only on test files the agent itself wrote, and the three other
ablated guards act on information the scorer only partially shares: the
claim-check and the scorer both test file existence, but the claim-check
runs on whatever the model claimed in its finish summary, and the
no-work and plan-completeness guards act on run-time information ---
productive tool use and the agent's own declared plan --- that the scorer
never sees (the published scoring is fully deterministic --- milestone
and final-artifact checks only; the scorer's optional LLM-judge component
was disabled for every published batch, so no judge enters any \lrab{}
number in this paper). The
per-task decomposition bounds how much of the effect this overlap can
explain: the net $-0.098$ is carried by four tasks at roughly $+0.71$
each (LH-01, WF-13, WF-04, WF-02) against three tasks the gate hurt
(WF-07 $-0.571$, WF-10 $-0.322$, WF-15 $-0.143$). Only one of the four
contributors (WF-04) stages a benchmark test file; its $+0.714$ is $41\%$
of the net delta. The remaining majority comes from tasks where no
benchmark test exists for the harness to run --- there the gate's benefit
is refusing finishes that claim artifacts which do not exist, not
executing the grader's checks, and the delta is measured with the textual
re-read active in both arms (see above). We still flag the WF-04/LH-03 overlap as a
threat to validity, and note the external contrast: on $\tau^2$-bench,
where the finish gate cannot execute any acceptance check (scoring is
database end-state plus judge), the harness margin persists at $+0.065$
to $+0.119$.

\paragraph{Mechanism activation and cost in the published matrix.}
The archived transcripts let us count how often each mechanism actually
fired across the 72 \system{} cells of the released matrix (counting
log markers; the extraction script and full CSVs are in
\texttt{benchmarks/mechanism\_log/}): delivery self-check (task-text
re-injection) 65 times, todo-sync refusals 43, summary-to-artifact
cross-checks 23, pytest-guard refusals 13, plan-completeness refusals 11,
and the no-work finish refusal \emph{zero} --- on these tasks the model
always did nominal work before finishing, so that guard contributes
nothing to the margins above, consistent with the per-task decomposition.
The anti-loop family intervened 95 times (mostly repeated-call
intercepts), plan-staleness nudges fired 99 times, the edit-safety guard
9 times, and format rescue 4 times. Tool-call volume per cell (median):
\system{} 33.5, goose 15.0, opencode 5.0, agent-mini 7.0 (transcript
formats differ across arms, so only magnitudes are comparable); \system{}
is the most tool-intensive arm, which is the intended behavior of a
harness that keeps a small model moving through concrete steps. Peak
prompt usage per \system{} turn (the archived context counters are the
backend's own \texttt{prompt\_eval\_count} readings) has median
14{,}624 tokens and maximum 32{,}767 --- the configured 32K budget ---
so even the heaviest cells finish inside the flat-prefill regime rather
than overflowing it. Wall-clock medians per cell run 440\,s (\system{}),
692\,s (goose), 450\,s (opencode) and 116\,s (agent-mini); only one of the
288 published cells was terminated by its budget (goose, WF-08), and the
per-arm table with turns, tool calls, token counters and the timeout
census is in \texttt{benchmarks/mechanism\_log/resource\_table\_2609.md}.

\begin{table}[t]
\centering
\small
\caption{Failure forms coded on all 288 published cells: number of cells
showing each form (a cell can show more than one). The upper block is
coded from recorded fields (\texttt{failure\_mode}, \texttt{exit\_code},
artifact-check details) and is comparable across arms; the lower block
counts harness log markers in the archived transcripts and has no
counterpart in the baselines (\emph{n/a}). Coding script, per-cell CSV and
the two-coder agreement check: \texttt{benchmarks/failure\_forms/}.}
\label{tab:failforms}
\begin{tabular}{lcccc}
\toprule
Failure form & \system{} & goose & opencode & agent-mini \\
\midrule
Platform exit (crash / error)                & 1 & 1 & 3 & 0 \\
Budget exhausted (timeout)                   & 0 & 1 & 0 & 0 \\
Provider disconnect                          & 0 & 0 & 0 & 8 \\
Never invoked a tool                         & 0 & 0$^{\dagger}$ & 21 & 0 \\
Stopped short (final score $<0.5$, no crash)& 4 & 23 & 32 & 44 \\
Incomplete delivery (a product check failed) & 26 & 44 & 53 & 61 \\
\midrule
Loop interceptions                           & 34 & \emph{n/a} & \emph{n/a} & \emph{n/a} \\
Malformed-argument rejections                & 3 & \emph{n/a} & \emph{n/a} & \emph{n/a} \\
Context-pressure events                      & 4 & \emph{n/a} & \emph{n/a} & \emph{n/a} \\
Finish refused (named artifact missing)      & 40 & \emph{n/a} & \emph{n/a} & \emph{n/a} \\
\bottomrule
\end{tabular}
\\[2pt]
{\footnotesize $^{\dagger}$one goose cell has no transcript (four
executions: two legitimate timeouts and two externally killed retries;
zero-scored by decision, Section~\ref{sec:protocol}), so
its count is undecidable.}
\end{table}

\paragraph{Failure forms across the matrix.}
Counting mechanism firings shows that the mechanisms are exercised;
coding what went wrong in each cell shows what they are exercised
\emph{against}. Table~\ref{tab:failforms} codes all 288 published cells.
The baselines' dominant form is not a wrong answer but an unattempted or
abandoned one: opencode never invoked a single tool in 21 of its 72 cells
(18 of them at 2B, consistent with its zero tool-call mean at that tier),
agent-mini lost 8 cells to backend disconnects that its own runner records
as \emph{completed}, and 23--44 cells per baseline arm stopped short of
the product checks with no crash and no timeout. In the \system{} arm the
behaviour that ends a baseline cell is instead an event the harness
intercepts: 40 cells contain a finish call the gate refused because a
named artifact was missing, and only 4 \system{} cells end below half
credit. These counts are incidence, not causal attribution --- in the
sampled cells three interceptions fired in cells that went on to full
credit, and the mechanism-activation counts above include the same
non-fatal firings. A second coder, reading a stratified 24-cell sample
independently, agreed with the extraction script on every sampled cell for
the record-derived and marker-derived forms (24/24 and 6/6 respectively);
for the single judgment form (\emph{stopped short}) the two agreed on
21/24, all three disagreements falling on the $0.5$ score boundary or on
the choice between the task-total and the final-stage score.

\paragraph{Development exposure of the mechanisms.}
Because the harness and the benchmark share authors, we disclose what each
ablated mechanism was developed against. The finish gate and the anti-loop
monitor were iterated \emph{with \lrab{} in the loop}: their v1.2.0 forms
were explicit responses to \lrab{} long-horizon failures (early-finish
abandonment; a same-tool-call rule that disarmed legitimate batch work),
and the WF-08 incident (156 calls, zero artifacts $\to$ 15 calls, three
artifacts, prefill byte-identical) drove the anti-loop escalation ladder.
The verification-feedback hint predates \lrab{} (developed against an
earlier internal benchmark, 2026-08-19..26, before \lrab{} existed on
2026-08-28), with \lrab{}-period strengthening after a WF-08 12B failure.
The flat prefill also predates \lrab{} (root cause: a Q\&A tool-demonstration
loop, 2026-08-20) and, as Section~\ref{sec:ablation} notes, is inert on
\lrab{}'s narrow tool surface. Consequently, the $-0.098$ finish-gate and
$-0.015$ anti-loop deltas should be read as in-sample effect estimates for
mechanisms co-developed with the task set --- which is why the ablation is
reported as directional throughout. Of the four ablated arms, three
carry some in-sample exposure (the finish gate and anti-loop monitor were
iterated with \lrab{} in the loop; the verification-feedback hint predates
\lrab{} but was strengthened during the \lrab{} period), and only the flat
prefill is both pre-\lrab{} and, by its near-zero delta, inert here; the
external $\tau^2$-bench comparison --- same code, no
benchmark-provided tests for the gate to run --- is the out-of-sample
anchor for the mechanisms the ablation cannot cleanly measure.

\begin{table}[t]
\centering
\caption{Leave-one-mechanism-out ablation (v1.5.0 code base, \lrab{} task
set, \texttt{gemma4:e2b}, $n=18$ per arm, one mechanism disabled per arm
via an environment gate). The baseline is the same-code full-mechanism arm;
these are the same 18 cells that form the 2B column of
Table~\ref{tab:matrix}, so the ablation and the headline results share one
batch and one code version. Deltas are exact per-task paired differences and
can differ in the last digit from the difference of the rounded totals
shown.}
\label{tab:ablation}
\begin{tabular}{@{}lcc@{}}
\toprule
Arm & Total & $\Delta$ vs.\ baseline \\
\midrule
full mechanism set (baseline) & 0.821 & --- \\
$-$ finish gate & 0.723 & $-0.098$ \\
$-$ verification feedback & 0.772 & $-0.048$ \\
$-$ anti-loop & 0.805 & $-0.015$ \\
$-$ flat prefill & 0.818 & $-0.003$ \\
\bottomrule
\end{tabular}
\end{table}

\begin{figure}[htbp]
\centering
\includegraphics[width=\linewidth]{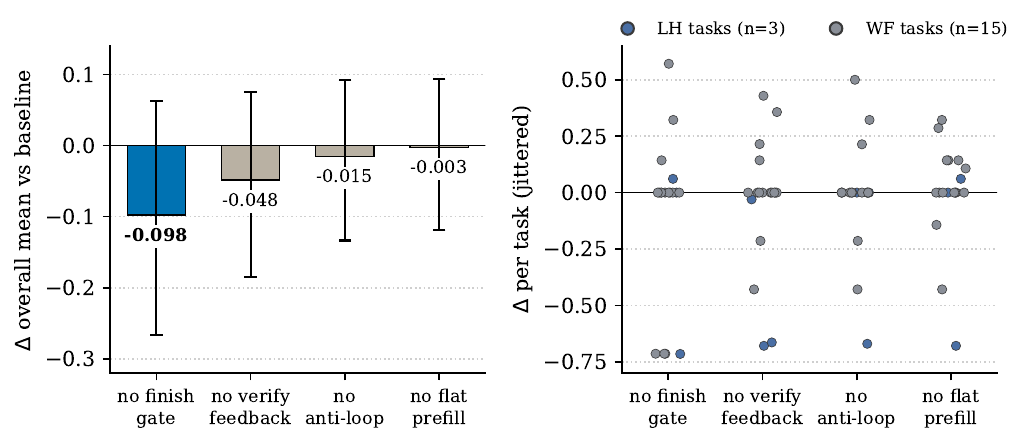}
\caption{Leave-one-mechanism-out ablation (v1.5.0 code base, \lrab{} task
set, \texttt{gemma4:e2b}, $n=18$ per arm; the baseline is the same-code
full-mechanism arm of Table~\ref{tab:ablation}). \emph{Left:} mean score
change per disabled mechanism with bootstrap 95\% confidence intervals over
10,000 task-level resamples (finish gate $[-0.266, +0.063]$, verification
feedback $[-0.185, +0.076]$, anti-loop $[-0.134, +0.092]$, flat prefill
$[-0.119, +0.093]$; every interval crosses zero, so the effects are
directional; the finish-gate bar alone is colored, marking the largest
single delta --- the coloring carries no separate statistical status).
\emph{Right:} per-task changes, colored by task
family (LH versus WF). Much of the LH-segment decline across the four arms
is carried by the common-cause shift of LH-01, which falls in every arm
(sensitivity in Section~\ref{sec:ablation}); under this design,
mechanism-level readings of the family split are not separable. Nor is
the WF family unchanged: disabling verification feedback leaves the WF
mean $+0.033$ (slightly better), and the finish-gate arm's WF mean is
$+0.074$.}
\label{fig:ablation}
\end{figure}

Table~\ref{tab:ablation} gives the totals, and Figure~\ref{fig:ablation}
decomposes the per-task changes by task family. No removal helps, and the
gradient is consistent with the design motivation of
Section~\ref{sec:mechanisms}, but it is directional rather than
statistically significant: task-level bootstrap 95\% confidence intervals
(10,000 resamples) are wide and all cross zero (finish gate $[-0.266,
+0.063]$, verification feedback $[-0.185, +0.076]$, anti-loop $[-0.134,
+0.092]$, flat prefill $[-0.119, +0.093]$). At $n=18$ per arm the design
cannot separate mechanism effects from task-level variance, so we read the
ordering rather than any single value: removing the finish gate reduces the
total by 0.098, the verification-feedback loop is next (a reduction of
0.048), and the anti-loop monitor contributes less at this scale
($-0.015$). Removing the flat prefill is close to neutral at 2B ($-0.003$),
and we interpret this near-neutral result as a property of the task set:
the mechanism's benefit is domain routing --- keeping each task's prefill
small when the total tool surface is large --- and tasks whose tool surface
is already narrow do not exercise it. We therefore do not claim every
mechanism is necessary on this evidence; the flat prefill is an enabler of
the byte-level budget guarantee (its cost is bounded by a unit test and CI)
rather than a score-maximizing one, and a harness stripped of the neutral
mechanisms is a reasonable variant for future work. The ablation covers one model at 2B
and 18 tasks per arm; it does not establish marginal contributions at 4B or
above (discussed in Section~\ref{sec:threats}).

\paragraph{Common-cause disclosure: LH-01.}
One per-task pattern must be disclosed before any mechanism is read out of
Table~\ref{tab:ablation}. LH-01 falls to 0.214--0.259 in \emph{all four}
ablation arms, from a baseline of 0.929: a drop of 0.670--0.715 whose
range across the four arms is only 0.045. Four different disabled
mechanisms, four independent executions, and the same task falls by the
same amount each time --- this does not behave like a mechanism effect but
like a common cause: LH-01 is a 180-minute multi-stage task, the most
sensitive task in the set to batch conditions, and it may itself be
bimodal. Excluding LH-01 from all five arms, the deltas become: finish
gate $-0.098 \to -0.062$ (the direction survives, 37\% smaller),
verification feedback $-0.048 \to -0.011$ (approaching zero), anti-loop
$-0.015 \to +0.023$ (sign flip), and flat prefill $-0.003 \to +0.037$
(sign flip). Only the finish gate survives in direction once the
common-cause task is excluded; the nominal anti-loop and flat-prefill
effects are entirely explained by that single cell.

The follow-up experiment settles it. We re-ran the full-mechanism arm
fresh on 2026-09-22 (v1.5.0 code, same model manifest, same budgets, the
same judge-free deterministic scoring as the ablation protocol) plus one
new arm, \texttt{finish\_gate\_text\_only}, which keeps only the
completion-time re-injection of the task text and disables the four
executable guards \emph{and} the todo-sync reminder. Two findings:
\begin{itemize}
  \item \textbf{The common cause is confirmed, and it is not limited to
  LH-01.} The fresh full-mechanism arm lands in the low mode by itself:
  LH-01 0.286 (matrix batch 0.929), LH-02 0.191 (0.878), LH-03 0.420
  (0.969); the 18-task mean is 0.740 against the matrix batch's 0.821,
  with single-cell swings on workflow tasks of up to $\pm 0.43$. With
  every mechanism enabled, the drop reproduces, so the LH-family movement
  in Table~\ref{tab:ablation} carries no mechanism-level information. The
  same reruns also fix the scale against which the deltas must be read:
  single-cell swings reach $\pm 0.5$ on long-horizon tasks, and --- the
  like-for-like comparator --- same-night replications of the
  \emph{same arm} move its 18-task mean by up to 0.069, the size of every
  nominal delta in the table. The four published deltas,
  including the finish gate's $-0.098$, are directional estimates from
  single trials, not measured marginal effects.
  \item \textbf{The one clean matched comparison favors the executable
  guards.} The two new arms ran in one batch on the same night, so their
  pairing is the only comparison in this section not separated by a batch
  gap. To move past single trials we repeated the matched pair three
  times (three independent same-night batch runs of each arm, 18 tasks
  each, published as \texttt{ablation\_reps\_2609.csv}). All three
  pairings favor the executable guards: total paired differences
  $+0.096$, $+0.133$, $+0.056$ (mean $+0.095$), workflow-segment
  pairings $+0.145$, $+0.157$, $+0.057$, and long-horizon $-0.148$,
  $+0.013$, $+0.051$ --- the long-horizon sign is not stable, consistent
  with that segment's variance (below). For scale, the same-configuration
  arms moved by a total-mean range of $0.069$ across their three trials
  (workflow means within $0.05$, per-task swings up to $0.75$ on
  long-horizon cells), so the matched-comparison estimate carries
  roughly that resolution: the executable guards are worth on the order
  of a tenth of a point over the textual re-read alone, same direction
  and magnitude as the original single-trial $-0.098$, now with three
  concordant trials behind the sign.
\end{itemize}
Per-cell scores for both new arms are published alongside the original
five (\texttt{benchmarks/ablation/}). The main matrix itself shares this
exposure: its long-horizon cells are single trials from the September
13--14 batch, and the same-night evidence above shows a fresh full-
mechanism execution can land $0.5$ lower on them; the LH-segment
magnitudes of Table~\ref{tab:matrix} should be read with that batch
variance in mind (the all-task and workflow-segment margins are the
stable quantities; see Threats).

\section{Threats to Validity}
\label{sec:threats}

\paragraph{Self-built benchmark.}
\lrab{}'s tasks and rubrics were authored by the same people who built
\system{}. We cannot rule out that the task set favors our harness's
mechanisms; indeed, it was designed to exercise the failure forms in
Section~\ref{sec:mechanisms}, which is the point of a controlled experiment
but also its bias. Mitigations include the full publication of tasks,
fixtures, scoring code, and raw cells, in the spirit of reproducible agent
evaluation \citep{agentsthatmatter}; the explicit framing as a controlled
experiment rather than a leaderboard; and the external $\tau^2$-bench
replication, whose benchmark, tasks, and scorer are third-party. The two
legs agree in direction.

\paragraph{External-benchmark coverage.}
$\tau^2$-bench is our only three-arm external comparison (a fourth CLI
arm, goose, is deferred for throughput), and it comes from
one family (one lab, three domains). This is not for lack of searching; it
reflects a structural constraint. A harness comparison requires the
\emph{harness} to be the experimental variable, which excludes every suite
that supplies its own agent loop, and it additionally requires the suite to
discriminate at local-model scale; dynamic GUI-device environments such as
AndroidWorld \citep{androidworld} add device emulation to that list and were
not candidates for the same reason. We examined five external candidates
against these two filters and excluded each for cause: AppWorld
\citep{appworld} (all-zero
at every local model tier we ran), ToolSandbox \citep{toolsandbox} and
AgentBoard (the
benchmark provides the agent loop; only the model varies), function-calling
emission suites (model-side emission with no environment and no multi-step
loop; e.g.\ JSON-schema emission benchmarks \citep{jsonschemabench}),
CRMArena \citep{crmarena} (we built a local backend against the official data
snapshot and probed it: gemma4:e2b 0/5, qwen3.5:4b 0/5, and the 35B tier
at 40+ minutes per task with the first completed task scored 0 ---
no discrimination at usable throughput), and GAIA (no discrimination for
local models in our runs). We report this filter explicitly so that the
single-family external evidence can be weighed for what it is; the
cross-benchmark agreement between \lrab{} and $\tau^2$ on the shared
harnesses is the strongest external check available under the constraint.

\paragraph{Development--evaluation overlap.} At least one mechanism (the
anti-loop gate, M4) was developed against a task that also appears in
\lrab{} (WF-08): its before/after evidence is that task. More broadly,
Section~\ref{sec:ablation} discloses that three of the four ablated
arms carry some in-sample exposure (finish gate, anti-loop monitor,
LRAB-period strengthening of verification feedback); only the flat
prefill is both pre-\lrab{} and inert here. There is no
held-out task split and no temporal separation between mechanism development
and evaluation, so test-set overfitting cannot be ruled out. We mitigate by
publishing every cell so the overlap is visible, and by the external
$\tau^2$ results, which were produced after the mechanisms were frozen; a
held-out task split remains future work.
\paragraph{Single machine and platform.}
All 288 cells ran on one Windows 11 machine (an integrated GPU, 32\,GB
shared memory). We do not know how the ranking transfers to Linux, to
discrete GPUs, or to different memory envelopes; agent-mini's 35B collapse
is specifically a Windows interaction and would presumably not reproduce on
POSIX shells. \system{} itself is Windows-first, which has two
consequences: the harness matches the benchmark platform exactly, and it is
untested elsewhere.

\paragraph{Campaign ordering (temporal confounding).}
The three baseline arms ran 2026-09-01..04 and were frozen at their stock
versions; the \system{} arm ran 2026-09-13..14 with the final v1.5.0 code;
the goose and opencode arms were re-shot 2026-09-18..20 under the unified
protocol. Execution order was blocked, not randomized, so machine state,
driver stack, and background load could in principle have drifted between
campaigns. Model weights are pinned by manifest digest, which controls the
model but not the runtime stack. Two facts partially compensate: the
goose/opencode re-shots cross the original campaign boundary and reproduce
the direction of the baseline campaign's results, and every cell is
published, making anomalies auditable. A fully interleaved replication
(round-robin task order across all four arms) remains the strongest check
and is future work.

\paragraph{Model set.}
Four models, one per size point, all open-weight and served by one backend.
Two further small-tier candidates were pulled and piloted for a second-family
check and rejected \emph{before} entering the matrix, on a criterion set in
advance: a candidate must form schema-valid tool calls. \texttt{minicpm5-2b}
and \texttt{llama3.2:3b} both fail that criterion in the same way --- they
emit argument keys the tools do not define (e.g.\ \texttt{\{"fn": \ldots\}}
where \texttt{path} is required), repeat the malformed call after the
harness's corrective message and worked example, and end with no artifacts
(the first made 175 rejected calls across 449 turns; the second scored 0.00,
0.06, 0.31 and 0.00 on its first four tasks, all without a single
schema-valid write). Their transcripts are archived with the data release.
This is a model-side floor that none of the harness compensations we
document repairs, and it is reported as such rather than as a harness
result. Our ``2B'' label in Table~\ref{tab:matrix} follows the Ollama tag naming:
\texttt{gemma4:e2b} is a 5B-total / 2.3B-active-per-token (MatFormer)
model, as stated in Table~\ref{tab:models}. The second small-tier family of
Section~\ref{sec:results} (\texttt{qwen3.5:2b}) is a dense 2.3B model run
through the same 18-task protocol, so the two low-tier results together
cover a MatFormer and a dense model at that parameter point. The matrix includes no proprietary model; the frontier probe of
Section~\ref{sec:probe} covers one hosted model, but with a single trial
per arm and a cloud endpoint that is not a frozen snapshot, so harness
effects at that scale remain indicative rather than established.

\paragraph{Scoring subjectivity residuals.}
Scoring is deterministic and artifact-based, but the artifacts-and-checks
rubrics were authored by us. Determinism removes run-to-run judgment
variance, not design-time rubric bias. Partial credit weights (e.g.\ a cell
scoring $1/7$ or $4/7$) reflect our judgment of task structure.

\paragraph{Single trial under greedy decoding.}
Every cell is a single run (one retry, latest attempt wins), with temperature
pinned to 0 for every arm. Greedy decoding removes sampling randomness at the
source, but it does not remove execution-level variance: the same-night
full-mechanism re-run of Section~\ref{sec:ablation} (identical code, model,
budgets, and scoring) moved long-horizon cells by $0.5$--$0.7$ and workflow
cells by up to $0.43$ against the matrix batch, and across three repeated
same-configuration trials the arm's total mean ranged over $0.069$ with
per-task long-horizon swings reaching $0.75$. Long-horizon magnitudes in
Table~\ref{tab:matrix} --- including \system{}'s LH-segment lead --- are
therefore batch-fragile and should be read as one draw per cell; the
all-task and workflow-segment margins are the stable quantities. The
task-level paired tests and bootstrap intervals of
Section~\ref{sec:results} are the inference this design supports, the
ablation results stay directional (their intervals reflect task variance,
not sampling), and the smaller segment margins (e.g.\ 12B: 0.906 vs.\
0.772) deserve replication.

\paragraph{Cloud user simulator.}
The $\tau^2$ comparison uses a cloud simulator (\texttt{qwen3.8-flash})
instead of the official gpt-4o user, so absolute scores are not comparable
to the leaderboard; only the between-arm difference is controlled. The
judge (\texttt{qwen3.8-max}) is likewise a cloud model. Both are accessed
by alias through one endpoint, and a cloud alias is not a frozen snapshot;
all arms were sampled in the same September 2026 window, but an independent
replication today could meet a drifted simulator. Simulator
behavior could also interact with harness style (e.g.\ tolerance for terse
agent replies).

\paragraph{Ablation coverage.}
Section~\ref{sec:ablation} removes mechanisms one at a time, but only on a
single 2B model over 18 tasks per arm, against a same-code control baseline,
and its task-level confidence intervals all cross zero. It does not cover
4B and larger models or the multi-model matrix, and with
one mechanism disabled at a time it says nothing about interactions between
mechanisms. Marginal contributions at other model scales remain unmeasured.

\paragraph{Safety evaluation scope.}
The five-ring evaluation (Section~\ref{sec:m14}) is one run per case over
54 probes and two models, on a probe set we wrote ourselves: it
establishes that the rings fire on realistic requests and characterizes
what leaks through, but the per-ring rates carry single-run noise, the
probe set has no external red-team provenance, and both confirmed leaks
are bash-side rule gaps documented as findings rather than silently
patched in the evaluated tree. The false-completion-claim phenomenon it
exposes is a model-level integrity problem that no host-side gate in the
evaluated version closes; we report its count and the per-case evidence
notes rather than claim a fix.

\section{Discussion and Future Work}
\label{sec:discussion}

\paragraph{What the results suggest.}
For small models, the harness is not peripheral glue but a substantial part
of the capability surface. The 2B segment shows this most clearly: same
model, same budgets, and one harness completes 0.821 of its cells while
others complete 0.017--0.271. The failure forms behind that gap (prefill
overflow, unbounded demonstration loops, silent abandonment, malformed calls
discarded by strict parsers) are deterministic engineering problems, and
none of them requires model retraining. The frontier probe sharpens the
attribution from the other direction (Section~\ref{sec:probe}): with the
model pinned near its ceiling, harness choice alone spans 0.478--0.997 on
the same tasks, so a weak scaffold strands capability that a stronger one
recovers. The ablation puts directional
numbers on this attribution at 2B: the largest single loss comes from
removing the finish gate (Section~\ref{sec:ablation}). We expect several
mechanisms (flat
prefill with a hard budget, finish gates, loop signatures) to transfer to
larger models with weaker budgets, though our data cover only up to the 35B
tier of Table~\ref{tab:models}.

\paragraph{Beyond Windows.}
\system{} is Windows-first; porting the harness (notably M7's encoding layer
and M8's path normalization) to macOS and Linux is planned. The agent-mini
result shows platform fit can dominate model capability in this size class,
which elevates cross-platform measurement beyond routine porting work.

\paragraph{A human baseline for \lrab{}.}
The natural missing baseline in \lrab{} is human performance: participants
completing the same 18 tasks under identical budgets. A careful human
baseline (a small number of participants, task order randomized, artifacts
scored by the same deterministic scorer where applicable) would calibrate
what 0.886 means; we consider this the most valuable single addition to the
benchmark.

\paragraph{Scaling the ablation.}
The leave-one-mechanism-out study of Section~\ref{sec:ablation} covers one
model at 2B and 18 tasks per arm. Extending it across the full matrix
(every mechanism at every model size) would turn the design narrative of
Section~\ref{sec:mechanisms} into complete marginal-contribution estimates.
The published matrix and open harness make this feasible for us or for
others.

\section{Conclusion}
\label{sec:conclusion}

We presented \system{}, a local-first agent harness for Windows and Ollama
that treats small-model failure forms as first-class engineering targets.
The evaluation is a fully published controlled experiment: 288 cells, four
harnesses, four open models (2B--35B), one machine, and deterministic
artifact-based scoring. \system{} reaches 0.886 overall against 0.631,
0.479, and 0.405 for three stock baselines, is the only harness without a
small-model cliff, and leads the long-horizon segment at 0.827 (single
draw per cell). An external
replication on $\tau^2$-bench in all three domains (0.856 vs.\ 0.791 and
0.737 over 278 tasks, same local 4B model, matched user simulator, with the
fourth CLI arm deferred for throughput on this benchmark)
supports the direction on a third-party benchmark, and a frontier-model
probe --- one hosted model, four harnesses, the same 18 tasks --- spans
0.997 to 0.478, showing that a defective scaffold buries more than half of
a frontier model while well-formed scaffolds stay within 0.072--0.12 of each
other. A
leave-one-mechanism-out ablation attributes the largest single share of the
margin to the finish gate, and the thinking configuration itself emerges as
a harness-level lever with task-dependent costs. \lrab{} is self-built
and carries its authors' bias; we publish every cell so that the results can
be independently recomputed or refuted. Our results suggest that a
substantial share of what is attributed to model incapacity in this setting
is in fact scaffold defect, and such defects are addressable.

\section*{Acknowledgments}
The authors used LLM-based assistants during code development and manuscript
editing. All experiment design, data collection, analysis, and the final
judgments expressed here are the authors' own.

\paragraph{Author contributions (CRediT).}
H.\ Wang: software, benchmark design, data curation, investigation,
visualization, validation, and writing (original draft and review \&
editing, jointly with T.\ Huang).
T.\ Huang: conceptualization, methodology, and writing (original draft and
review \& editing, jointly with H.\ Wang).

\paragraph{Conflict of interest.}
T.\ Huang is an employee of Honor Device Co., Ltd. This work is
independent of that employer: it was conducted on personal equipment, uses
no employer resources, and the employer had no role in the study design,
analysis, decision to publish, or preparation of the manuscript.
H.\ Wang declares no conflicts of interest.

\appendix
\section{Case studies: the mechanisms at work}
\label{app:cases}

The aggregate tables compress 288 runs into means; this appendix shows
three exchanges, verbatim from the published transcripts, where a mechanism
 visibly changes what happened. (Transcripts log both the Chinese harness
messages and the model's English output; we gloss the markers.)

\paragraph{C1: format-leak salvage (WF-01, 2B, turn 20).}
The model completed the literature-review task and tried to finish, but
emitted the \texttt{finish} call as plain text instead of a structured
call. The parser recognized the call's structure inside the prose, salvaged
it into a real invocation, and the run ended as a scored completion:

\begin{quote}
\small\ttfamily
[20] \ lightning-mark: salvaged plain-text tool call: finish\\
{[}\dots{]} all requirements satisfied, task complete.\\
{[}20{]} finish \{``summary'': ``Completed the mini literature review \dots''\}\\
===== TASK COMPLETE =====
\end{quote}

Without salvage, this turn would have been an empty round: the same cell's
score depends on the artifacts already written, but the completion claim
itself would have been lost in prose.

\paragraph{C2: the finish gate (WF-01, 12B, turns 17--18).}
The model declared completion after checking off its plan. The gate
re-injected the original task text (1{,}648 characters) and required a
self-check before accepting the finish. The model then re-finished with
the concrete deliverables named:

\begin{quote}
\small\ttfamily
{[}17{]} finish \{``summary'': ``Completed the mini literature review \dots''\}\\
{[}17{]} warn: delivery self-check --- task text re-injected, re-finish
required\\
{[}18{]} finish \{summary naming \texttt{shortlist.md},
\texttt{data\_table.md}, \texttt{contradictions.md}, the seven selected
papers, and the documented data discrepancies\}
\end{quote}

The ablation shows the population-level version of this exchange: removing
the gate costs a nominal $0.098$ overall, the largest single-mechanism
delta in the directional ablation (Section~\ref{sec:ablation}).

\paragraph{C3: the anti-loop guard (WF-08, 4B, before/after).}
The signature-level guard was developed against this task, where the stock
loop degenerated into 156 identical tool calls with zero artifacts in the
target directory. After the fix --- same code, same task, same model, with
the static prefill byte-identical (asserted in CI) --- the same cell runs
15 tool calls and delivers 3 artifacts. The guard's interception message
tells the model why the repeat cannot succeed and what to do instead
(``read the actual state first, or change strategy''), converting a silent
spin into a redirect. The $\tau^2$ adapter carries the same idea
(same-signature repeat $\geq$ 4 triggers a corrective nudge).

\section{Reproducibility}
\label{app:repro}
Everything needed to re-derive the numbers in this paper is public:

\begin{itemize}
  \item \textbf{Repository.} \url{https://github.com/Mingbird/Mingbird-agent}
  (Apache-2.0). The data release of record is commit \texttt{61fc6aa} on
  \texttt{main} (benchmarks CSVs, unified $\tau^2$ manifests, ablation scores,
  significance script); the numbers first appeared in \texttt{7c99941}.
  \item \textbf{Arm versions.} All baselines stock and unpatched:
  goose 1.48.0, opencode 1.18.23, agent-mini 0.3.1 (each installed
  2026-08-27/28, before the 09-01..04 freeze, and unchanged since ---
  file timestamps verify no binary changed across the 09-18..20 re-shot);
  \system{} arm: v1.5.0 tag (\texttt{1ee92d1}); backend: Ollama 0.33.2
  (installed 2026-08-28; the executable is unchanged across all collection
  windows, and a 2026-09-13 commit in the repository records the same
  version empirically).
  \item \textbf{Results vs.\ versions.} The \lrab{} matrix, the ablation
  (including the 2026-09-22 repeats), and the $\tau^2$ arm all ran on the
  v1.5.0 tag (\texttt{1ee92d1}; factory prefill 774 tokens, pinned by that
  tag's unit test); the $\tau^2$ arm in particular drives a frozen v1.5.0
  tree bundled with its adapter. Two components ran later, on the product
  repository rather than the frozen tag, and are disclosed as such:
  the frontier-model probe of Section~\ref{sec:probe} drives the product
  repository's working tree as of its run date --- the 2026-09-20 four-arm
  campaign on the post-v1.7.0 tree (the v1.8.0 commit landed 17 minutes
  after the probe's first cell), and the 2026-09-22 mechanism-ablation
  re-run on the v1.8.2-era tree --- so probe-versus-matrix comparisons
  (e.g.\ the probe/4B juxtaposition in Section~\ref{sec:probe}) are
  cross-version for the \system{} arm; and the static prefill cost curve
  of Section~\ref{sec:mechanisms} was measured on the same post-v1.7.0
  tree (2026-09-20). The five-ring safety evaluation of
  Section~\ref{sec:m14} likewise ran on the v1.8.2-era tree rather than
  the v1.5.0 tag: the v1.5.0 tag predates two of the five rings. Ring
  history by first release, verified by matching every ring's refusal
  markers against the repository's release history: irreversible-command refusal and
  the workspace path boundary entered at v1.1.0, the sub-agent sandbox
  at v1.3.0, and risk-graded system mutation and reversibility at
  v1.6.0 --- no tag before v1.6.0 carries the full five-ring model, so
  the evaluation ran on the working tree of its run date. The second
  small-tier family of Section~\ref{sec:results} (\texttt{qwen3.5:2b}) ran
  on the same v1.5.0 code base as the matrix, with one protocol deviation:
  cells were run back-to-back without the per-cell backend restart used for
  the matrix campaigns, because that restart path proved unreliable on the
  machine on the day (a killed backend was repeatedly judged unreachable
  while other processes were using the same local Ollama instance). Scores
  are unaffected --- scoring is deterministic and artifact-based, and no
  cell hit its budget --- but the row's wall-clock is not comparable to the
  matrix's. The paper
  documents the harness at v1.6.0 (prefill 797 tokens = 774 + a 23-token
  safety-pad wording change, both pinned by unit test; 441 tests as
  collected by pytest, including parametrized cases; the release of the
  evaluation date, v1.8.2, collects 461 --- 435 unit + 26 integration ---
  with the count reconciliation maintained in the repository README). The
  v1.5.0$\to$v1.6.0
  delta between the measured code and the documented baseline is the
  safety cushion that completes the five-ring model (rings 3 and 5), which
  adds no mechanism evaluated in the ablation
  and whose static cost is the disclosed 23 tokens.
  \item \textbf{Models.} The four model families are public releases
  (Table~\ref{tab:models}): gemma4 by Google, qwen3.5 by Alibaba, and
  Ornith-1.5-35B-A3B by Ornith AI \citep{ornith15}. All four tags served as
  Q4\_K\_M ggml quantization. Ollama manifest digests (SHA-256; tags are
  content-addressed, so a digest match certifies identical weights across
  all collection campaigns):
    \sloppy
    \begin{itemize}
      \item \texttt{gemma4:e2b}
        \texttt{7fbdbf8f5e45a75b\allowbreak{}b122155ed546e765\allowbreak{}b4d9c53a1285f62f\allowbreak{}d9f506baa1c5a47e}
      \item \texttt{qwen3.5:4b}
        \texttt{2a654d98e6fba55d\allowbreak{}452b7043684e9b57\allowbreak{}a947e393bbffa624\allowbreak{}85a7aac05ee4eefd}
      \item \texttt{gemma4:12b}
        \texttt{4eb23ef187e2c546\allowbreak{}2566d6a1d3bbbc2f\allowbreak{}1346d0b4327cbb66\allowbreak{}d58fffbcc9b2b05c}
      \item \texttt{ornith-1.5:35b}
        \texttt{9f3b89b2521908dd\allowbreak{}2e6f7a11fa368e62\allowbreak{}c8f89e1075f22604\allowbreak{}e4d1a76dd1240fcc}
        \\
        (each is the SHA-256 of the model's Ollama manifest; the full
        per-model record --- manifest hash, model-layer blob digest and
        size, projector layer where present, and the Ollama version ---
        is pinned in \texttt{benchmarks/models.lock} in the repository)
    \end{itemize}
  \item \textbf{\lrab{}.} Task definitions, fixtures, the deterministic
  scorer, and all 288 per-cell results (\texttt{benchmarks/lrab\_scores.csv}:
  harness, task, model, score, wall time, attempt directory). The matrix
  spans the three collection windows of Table~\ref{tab:campaigns}
  (agent-mini 09-01..04; \system{} 09-13..14 on the released code;
  goose/opencode unified-protocol re-shot 09-18..20; same machine,
  budgets, models, and scorer throughout, Section~\ref{sec:lrab}).
  \item \textbf{Statistics.} \texttt{analyze\_significance.py} in the
  repository reproduces every Wilcoxon test and bootstrap interval
  (10,000 task-level resamples, fixed seed) from the CSV alone.
  \item \textbf{$\tau^2$-bench.} Per-trial manifests for all three domains
  and all published arms (the deferred goose arm excepted), including
  status, reward, and wall time
  (\texttt{benchmarks/tau2/}). Agent model: local
  \texttt{qwen3.5:4b} via Ollama; user simulator \texttt{qwen3.8-flash} and
  natural-language judge \texttt{qwen3.8-max}, both served through the
  DashScope (Alibaba Cloud) endpoint, identical across arms and sampled in
  the same September 2026 window. The $\tau^2$-bench release used is
  frozen with the repository's data release of record; the native arm is
  the upstream default LLM-agent implementation shipped with the
  $\tau^2$-bench release used, unmodified.
  \item \textbf{Ablation.} Per-cell scores for all five arms
  (\texttt{benchmarks/ablation/ablation\_scores.csv}); the baseline arm is
  the same 18 cells as the 2B column of Table~\ref{tab:matrix}.
  \item \textbf{Mechanism log and per-cell cost.} Per-cell mechanism
  activation counts (finish-gate family, anti-loop, plan nudges, edit
  guard, format rescue), tool-call counts for all four arms, and
  \system{} prompt-token peaks
  (\texttt{benchmarks/mechanism\_log/}, with the extraction script).
\end{itemize}

Hardware: integrated GPU, 32\,GB shared memory, Windows 11.
Budgets: 90 minutes per workflow cell, 180 minutes per long-horizon cell;
one retry, latest attempt wins; timeouts score 0.

\bibliographystyle{unsrtnat}
\bibliography{refs}

\end{document}